\pdfoutput=1

\documentclass[11pt]{article}
\usepackage{fancyvrb,cprotect}
\usepackage{hyperref}
\usepackage{url}
\usepackage{amsmath}        
\usepackage{amsfonts}       
\usepackage{amssymb}        
\usepackage{mathrsfs}       
\usepackage{bm}             
\usepackage{amsthm}         
\usepackage{graphicx}
\usepackage{multirow}
\usepackage{subcaption}
\usepackage{graphicx}
\usepackage{microtype}
\usepackage{multirow}
\usepackage{footmisc}
\usepackage{soul}
\usepackage{arydshln}

\usepackage{algpseudocode}
\usepackage{xspace}

\usepackage{times}
\usepackage{latexsym}
\usepackage{graphics}
\usepackage{graphicx}
\usepackage{color}
\usepackage{colortbl}
\usepackage{amssymb}
\usepackage{amsmath}
\usepackage{bbm}
\usepackage{booktabs}
\usepackage{arydshln} 
\usepackage[most]{tcolorbox}

\usepackage{xcolor}
\definecolor{myorange}{RGB}{2, 142, 2}
\usepackage[linesnumbered,ruled,vlined]{algorithm2e}
\usepackage[]{latex/acl}

\usepackage{times}
\usepackage{latexsym}

\usepackage[T1]{fontenc}

\usepackage[utf8]{inputenc}
\usepackage{hyperref}
\usepackage{microtype}

\usepackage{inconsolata}

\title{MuggleMath: Assessing the Impact of Query and Response Augmentation on Math Reasoning}

\author{Chengpeng Li\textsuperscript{12}\thanks{Work done during internships at Alibaba Group.} , Zheng Yuan\textsuperscript{2} ,
Hongyi Yuan\textsuperscript{2}$^*$,
Guanting Dong\textsuperscript{2}$^*$, Keming Lu\textsuperscript{2} \\
\textbf{Jiancan Wu\textsuperscript{1}, Chuanqi Tan\textsuperscript{2}, Xiang Wang\textsuperscript{1}\thanks{Corresponding author}\thanks{Xiang Wang is also affiliated with Institute of Dataspace, Hefei Comprehensive National Science Center.}, Chang Zhou\textsuperscript{2}} \\
\textsuperscript{1}University of Science and Technology of China \\
\textsuperscript{2}Alibaba Group \\
\texttt{\{lichengpeng.lcp,yuanzheng.yuanzhen,ericzhou.zc\}@alibaba-inc.com} \\
\texttt{\{wujcan,xiangwang1223\}@gmail.com}
}

\begin{document}
\maketitle
\begin{abstract}

In math reasoning with large language models (LLMs), fine-tuning data augmentation by query evolution and diverse reasoning paths is empirically verified effective, profoundly narrowing the gap between open-sourced LLMs and cutting-edge proprietary LLMs. 
In this paper, we conduct an investigation for such data augmentation in math reasoning and are intended to answer: 
(1) What strategies of data augmentation are more effective; 
(2) What is the scaling relationship between the amount of augmented data and model performance; and 
(3) Can data augmentation incentivize generalization to out-of-domain mathematical reasoning tasks?
To this end, we create two new dataset AugGSM8K and AugMATH, by complicating and diversifying the queries and sampling multiple reasoning paths from GSM8K and MATH.
We obtained a series of LLMs called MuggleMath by fine-tuning LLaMA models on AugGSM8K and AugMATH. MuggleMath substantially achieves new state-of-the-art on GSM8K and MATH.
A log-linear relationship and a segmented log-linear are presented between MuggleMath's performance and the amount of augmented data on GSM8K and MATH, respectively.
We also find that it is weak in out-of-domain math reasoning generalization from AugGSM8K to MATH and from AugMATH to GSM8K, which suggests that augmenting queries that cover a broader range of subjects is more beneficial for generalization. We release our codes and augmented
data in \url{https://github.com/OFA-Sys/gsm8k-ScRel}.


\end{abstract}

\section{Introduction}
The emergence of large language models (LLMs)~\citep{instructgpt,anil2023palm,gpt4} has profoundly revolutionized the field of natural language processing, exhibiting versatile performance in various tasks like code generation~\citep{chen2021codex,wizardcoder,wang2024dolphcoder,wei2023magicoder}, instruction following~\citep{longpre2023flan,zhou2023instructionfollowing,lei2024instructerc},
long context question answering~\citep{tworkowski2023focused,luo2024chatkbqa},
and math reasoning~\citep{cot,taylor2022galactica,lewkowycz2022solving}. 
Math reasoning as a representative reasoning task is widely studied to access the reasoning abilities in LLMs~\citep{gsm8k,hendrycks2021measuring}. 
Proprietary LLMs, such as GPT-3.5, and GPT4~\citep{gpt4} have shown exceptional mathematical reasoning abilities, while there remains a substantial gap between open-source LLMs, such as GPT-J~\citep{gpt-j} and LLaMA~\citep{llama,llama2}) and the cutting-edge proprietary models.

To enable better mathematical reasoning abilities in open-sourced LLMs, they generally undergo a fine-tuning stage on supervised reasoning datasets.
A series of efforts are committed to enhancing the mathematical reasoning capabilities of open-source LLMs, where a mainstream approach involves first augmenting new mathematical problems and answers, followed by supervised fine-tuning on the augmented dataset~\citep{rft,luo2023wizardmath,yu2023metamath}. 
This type of approach has achieved good results, and in this paper, we would like to explore what are the key factors affecting the effectiveness of data augmentation for mathematical reasoning tasks and the scaling relationship between the amount of data augmentation and model performance. 
Specifically, with the help of proprietary models (GPT-3.5 and GPT-4), we applied five types of mathematical problem augmentation methods based on human experience in creating variations of mathematical problems similar to \citet{wizardcoder,luo2023wizardmath}. 
We further generated multiple reasoning paths for each augmented problem since distinct reasoning paths can also enhance chain-of-thought reasoning~\citep{huang2022large,core,rft}. We obtained two new datasets called AugGSM8K and AugMATH after data augmentation on two widely used mathematical reasoning datasets GSM8K~\citep{gsm8k} and MATH~\citep{MATH}. By supervised fine-tuning on the open-source LLaMA~\citep{llama} and LLaMA-2~\citep{llama2} LLMs on AugGSM8K and AugMATH, we obtained a series of models dubbed MuggleMath. We find that with sufficient amounts of data, MuggleMath achieves a new state-of-the-art on GSM8K and MATH. In addition to this, we find a log-linear relationship between the performance of MuggleMath and the amount of data augmentation over a range of data volumes on GSM8K and a segmented log-linear relationship on MATH. 


Although MuggleMath achieves strong performance on the GSM8K and MATH test set, the rationales for performance improvement by data augmentation remain unclear.
We are therefore interested in the specific reason behind the performance improvement and whether it brings enhancement in LLMs' mathematical reasoning capabilities generally.

To validate the generalization of MuggleMath, we conduct multi-task learning and analyze the transferability with AugGSM8K and AugMATH.
We found that LLMs trained with supervised learning after data augmentation on GSM8K only bring marginal improvements to performance on MATH and the similar conclusion is fit for AugMATH and GSM8K.
By visualizing the data distribution in the embedding space of LLaMA-2-7B, we observe that the embedding distribution of problems in AugGSM8K is very close to that of GSM8K, but significantly different from the problem distribution in the MATH dataset.
The reason behind can be attributed to the fact that GSM8K and MATH have different reasoning difficulty, response style and require different mathematical knowledge.

The main contributions of our work can be summarized as follows:
\begin{itemize}
    \item By augmenting GSM8K and MATH with various queries and multiple reasoning paths, we curate GSM8K and MATH to two new datasets named AugGSM8K and AugMATH.
    \item We utilize AugGSM8K and AugMATH for fine-tuning the LLaMA and LLaMA-2 models to obtain MuggleMath, which greatly improves the in-domain performance of the open-sourced LLMs on GSM8K and MATH, achieving new state-of-the-art performances.
    \item We find a log-linear relationship between the accuracy of the model on the test set and the amount of data augmentation within a certain range while the coefficient is similar to augmenting new human-written samples on GSM8K. When it comes to MATH, a a segmented log-linear relationship is found. 
    \item We demonstrate that the performance gains from data augmentation on GSM8K and MATH are difficult to generalize to each other, which indicates a need of diverse original queries in augmenting math data.
\end{itemize}

\section{Related Works}

\paragraph{Mathematical Reasoning for Large Language Models}
Mathematical reasoning is a crucial ability to examine large language models~\citep{gsm8k,hendrycks2021measuring,cot,math401}.
The mathematical reasoning ability of LLMs can be enhanced by math-related pre-training~\citep{hendrycks2021measuring,lewkowycz2022solving,taylor2022galactica,lightman2023lets} and math-related supervised fine-tuning~\citep{rft,luo2023wizardmath,yue2023mammoth,yu2023metamath}.
Query augmentation~\citep{luo2023wizardmath,yu2023metamath} and response augmentation~\citep{huang2022large,zelikman2022star,ni2023learning,zhu-etal-2023-solving,rft} are useful techniques to improve math in-domain performances during SFT.
Query augmentation methods usually generate rephrased, easier, or harder problems and use proprietary LLMs to generate answers.
Response augmentation methods generate new reasoning paths for problems in the training set. They could rely on answers in the training set to filter the generated reasoning paths.
\citeauthor{rft} invests the scaling relationship on supervised LLMs math performance with pre-train loss, supervised data amount, and augmented reasoning path amount. 
Our work is further investigate on scaling relationships with query augmentation and out-of-domain generalization. 
MetaMath\citep{yu2023metamath} is a contemporary work that is similar to us in the augmentation method. The distinction lies in MetaMath's focus on rewriting original questions to create new ones using the questions' mathematical relationships. In contrast, our efforts are centered on generating new problems with equal or greater difficulty levels. 
Moreover, our work investigates the quantitative relationship between query and response augment amounts and in-domain and out-of-domain performances.

\paragraph{Data Augmentation for LLM}
Data augmentation is a common technique to improve downstream task performance in NLP~\citep{feng-etal-2021-survey}.
In the era of large language models, data augmentation is usually used for generating instruction following SFT datasets~\citep{wang-etal-2023-self-instruct,alpaca,xue2023occuquest}.
Queries~\citep{ultralm,wizardlm} and responses~\citep{orca} of SFT datasets can both be augmented by prompting state-of-the-art proprietary LLMs.
Compared with their work, we are concentrated on augmenting math SFT dataset and we are more interested in scaling relationships on in-domain and out-of-domain generalizations.

\paragraph{Out-of-Distribution Generalization} The challenge of out-of-distribution (OOD) generalization has garnered widespread attention across various domains~\citep{OOD1,OOD2,song2023large,OOD3,peng2024energy} in machine learning. This issue arises when the distribution of data encountered by a model during testing diverges from that of the training phase, leading to a decline in model performance. The OOD problem is multifaceted~\citep{OODP1,OODP2,OODP3,OODP4}, with subcategories such as covariate shifts and concept shifts, among others. To mitigate the effects of OOD scenarios, a diverse array of strategies has been developed, including unsupervised domain generalization~\citep{OOD2,OOD3}, stable learning~\citep{stablelearning,sd}, invariant representation learning~\citep{IL}, causal learning~\citep{causal}, and invariant risk minimization~\citep{IRM}, multi-task Learning \citep{dong2024abilities,wang2024multi} and contrastive learning~\cite{peng2023came}. Recent trends in the community have shown a growing preference for performance enhancement through data augmentation during the Self-supervised Fine-tuning (SFT) stage in large-scale models. However, the extent of OOD issues associated with this method and their severity remain underexplored. This study aims to fill this gap by conducting empirical experiments and providing a visual analysis to assess the impact of data augmentation on OOD generalization in the context of large models.
\section{Experiments}
We first introduce our experimental setup in Section \ref{sec:setup} and dataset augmentation method in Section \ref{sec:augment}. Then we conduct analyses spanning several aspects of data augmentation to answer the three research questions in abstract. For space saving, we mainly analyze the augmentation on GSM8K and detailed discussion of augmentation on MATH are list in Appendix \ref{app:AugMATH}.

\subsection{Experimental Setup}\label{sec:setup}

\paragraph{Problem Definition}
We define the math reasoning SFT dataset as $\mathcal{D}=\{q_i,a_i\}_i$, where $q_i$ is a question and $a_i$ is a reasoning path with an answer.
We augment the SFT dataset to a new dataset $\mathcal{D}'=\{q'_i,a'_i\}_i$. 
We apply SFT on the pre-trained language models and measure the augmented SFT dataset based on the accuracy of the in-domain test set $\mathcal{D}_{in}$ and out-of-domain test set $\mathcal{D}_{out}$.
We calculate the accuracy based on greedy decoding.

\paragraph{Datasets}
GSM8K~\citep{gsm8k} is a dataset with elementary school math word problems with 7,473 training problems and 1,319 testing problems. 
The test set of GSM8K is viewed as in-domain test datset for AugGSM8K and out-of-domain test dataset for AugMATH. MATH~\citep{hendrycks2021measuring} is a dataset with challenging high-school math problems. Problems are classified into the following topics: Prealgebra, Algebra, Number Theory, Counting and Probability, Geometry, Intermediate Algebra, and Precalculus. Problems in MATH are harder and more diverse than in GSM8K.
We use test set of MATH as our in-domain test dataset for AugMATH and out-of-domain test dataset for AugGSM8K.


\paragraph{Training} We employ state-of-the-art open-source LLMs for fine-tuning, including LLaMA-1 7B~\citep{llama}, LLaMA-2 7B, LLaMA-2 13B, and LLaMA-2 70B~\citep{llama2}, all of which undergo full fine-tuning. We adopt system prompt from \cite{alpaca} for fine-tuning and listed in Appendix \ref{app:alpaca}.
We use AdamW for optimization. The training proceeds for three epochs with a learning rate of 2e-5, a warmup ratio of 0.03, and a cosine learning rate scheduler. We do not apply early stops to choose checkpoints. The hardware setup involves 32 NVIDIA A100 GPUs. 

\subsection{Dataset Augmentation}\label{sec:augment}

\paragraph{Query Augmentation}
To generate new queries for GSM8K, we use \texttt{gpt-3.5-turbo-0613} and \texttt{gpt-4-0613} as the generator.
Inspired by Evol-Instruct~\citep{wizardcoder,luo2023wizardmath}, we find that the diversity and complexity of queries in augmented datasets play a vital role in improving math reasoning benchmark performance. 
We employ human  knowledge from authors in modifying mathematical problems for query augmentation. Below are five query augmentation methods used in our experiments: \textbf{Change specific numbers; Introduce fractions or percentages; Combine multiple concepts; Include a conditional statement; Increase the complexity of the problem.} 
The examples and detailed prompts we used for query augmentation are listed in Appendix \ref{app:prompt}. The examples of augmented queries are shown in Table \ref{Multi-query}.

\paragraph{Response Augmentation} Instead of using the trained SFT model proposed by \citeauthor{rft}, we use \texttt{gpt-3.5-turbo-0613} and \texttt{gpt-4-0613} to augment more reasoning paths .
The main reason is that we can not filter out wrong reasoning paths without final answers. 
Thus we need to use a model that is as accurate as possible which is the state-of-the-art LLMs \textit{ChatGPT}.   
We use a 1-shot prompt to ensure augmented response formats.
The response prompt we used for query augmentation is listed in Appendix \ref{app:response}.
Augmented responses can result in some unconventional answers, such as excessively long reasoning paths and reasoning paths that do not contain an answer at their end. We devise manual rules to filter out these corresponding query-response pairs and manual rules are detailed in Appendix \ref{app:response filter}.
The examples of augmented responses are shown in Table \ref{Multi-response}.



\paragraph{Augmented Dataset}
The original GSM8K training set has 7,473 samples. 
We augment 5 more queries for each query in the training set and yield $7,473 \times 5 = 37,365$ augmented queries.
We run this query augmentation three times with $\mathcal{D}_1, \mathcal{D}_3$ by GPT-3.5 and $\mathcal{D}_2$ by GPT-4, and $\|\mathcal{D}_i\|=37,365$.
Then we generate one response for each augmented query for $\mathcal{D}_i$ and apply response filtering. We 
consider the query-response pairs after filtering as $\mathcal{D}_i^j$.
We obtain approximately 30,000 query-response pairs for each $\mathcal{D}_i^j$.
To explore the performance differences of different augmented settings, we generate five responses on the augmented queries $\mathcal{D}_1$ with GPT-4’s temperature set to 1.0 ($\mathcal{D}_1^1\sim \mathcal{D}_1^5$), one response with GPT-4’s temperature set to 0.0 ($ \mathcal{D}_1^6$), and one response with GPT-3.5’s temperature set to 1.0 ($ \mathcal{D}_1^7$).
We also try a zero-shot response generation named $\mathcal{D}_1^8$.
We use GPT-4 to augment responses as $\mathcal{D}_2^1, \mathcal{D}_3^1$.
Since $\mathcal{D}_2^1$ is significantly larger than other subsets, we downsample it to $\mathcal{\hat{D}}_2^1$. We refer to the union of all augmented data and the original GSM8K training set as AugGSM8K, upon which we conduct experiments using various subsets.
Detailed augmented dataset notations are listed in Table \ref{tab:aug}.

\begin{table}[h]
    \small
    \centering
    \begin{tabular}{lccccc}
    \toprule
    \textbf{Subset} & \textbf{Query}  & \textbf{Response }& \textbf{Temp.} & \textbf{Size~(K)}  \\
    \midrule
    $\mathcal{D}$ & -& - & - & 7.5 \\
    $\mathcal{D}_1^1 \sim \mathcal{D}_1^5$ &GPT-3.5 & GPT-4 & 1 &30\\
    $\mathcal{D}_1^6$ &GPT-3.5 & GPT-4& 0 &30\\
    $\mathcal{D}_1^7$ &GPT-3.5 & GPT-3.5& 1 &25\\
    $\mathcal{D}_1^8$ &GPT-3.5 & GPT-4& 1  &30\\
    $\mathcal{D}_2^1$ &GPT-4 & GPT-4& 0 &35\\
    $\mathcal{\hat{D}}_2^1$&GPT-4 & GPT-4& 0 &30\\
    $\mathcal{D}_3^1$ &GPT-3.5 & GPT-4& 1 &30\\
    \bottomrule
    \end{tabular}
    \caption{The description of different subsets of the augmented in-domain dataset AugGSM8K.}
    \label{tab:aug}
\end{table}




\begin{figure}[t]
    \centering
    \includegraphics[width=\linewidth]{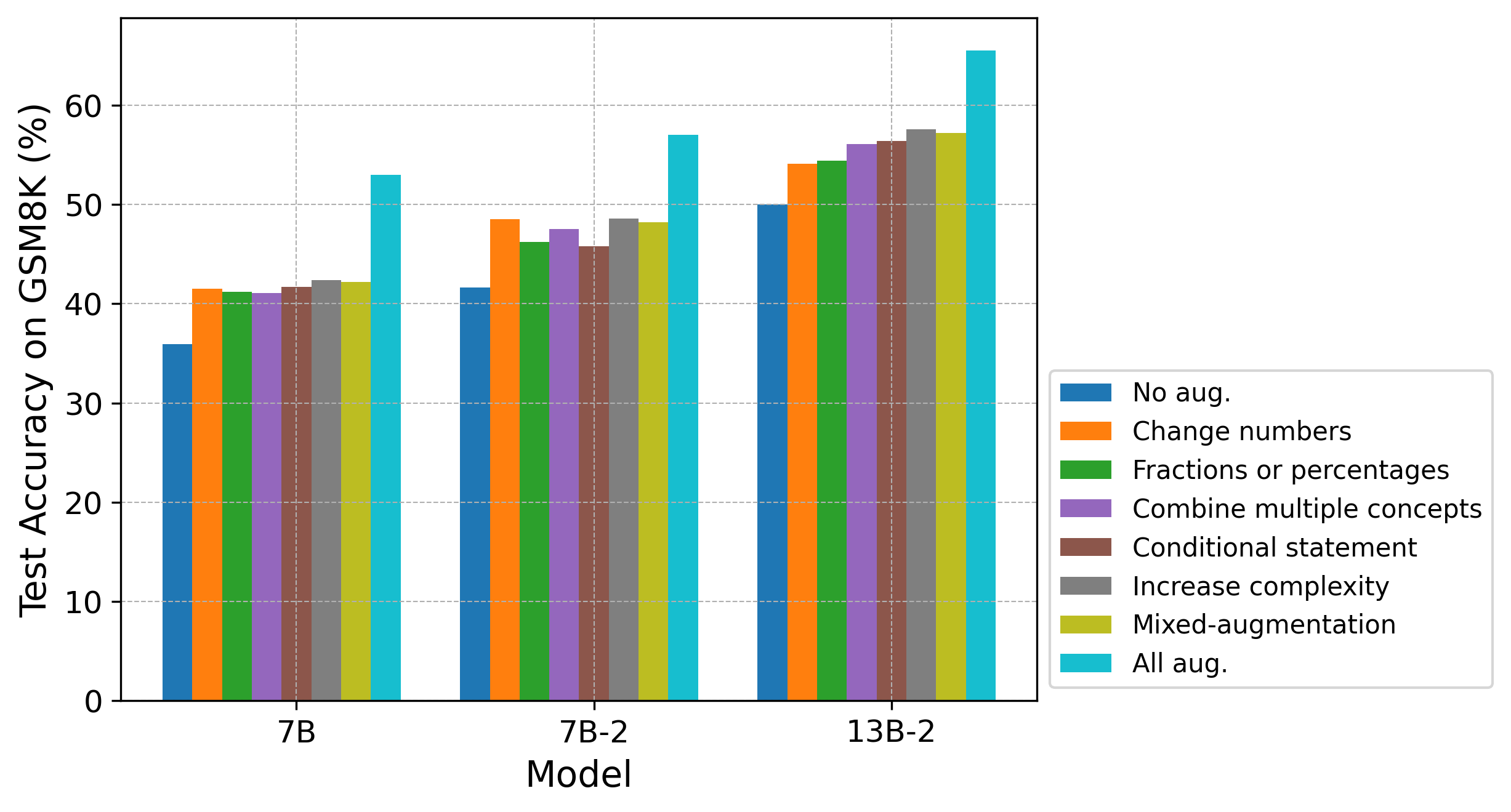}
    \caption{Comparison of test set accuracy on GSM8K for models of varying scales after fine-tuning on AugGSM8K subsets with different query augmentation strategies.}
    \label{query type}
\end{figure}


\subsection{RQ1. What Strategies of Data Augmentation are More Effective}\label{sec:in-domain}

\paragraph{Query Augmentation Types}
We want to examine whether query augmentation works for math reasoning SFT since \cite{luo2023wizardmath} applies PPO which cannot make an apple-to-apple comparison. 
Each query in the original training dataset is augmented with 5 different types. We cluster these queries based on the query types. We apply SFT on the original training set ($\mathcal{D}$), each query type augmentation, and a combination of them ($\mathcal{D} + \mathcal{D}_1^1$). Results are shown in Figure \ref{query type} and Table \ref{tab:query type}, where mixed-augmentation represents 1/5 of the augmented data of each strategy being randomly sampled and then combined.
Compared with no augment, each query augment method can improve the in-domain performance. \textbf{Increase complexity} augmentation method improves most among all of them. This suggests that enhancing the complexity of queries is one of the key factors influencing the sample efficiency of data augmentation. We randomly sampled one-fifth of the combined dataset and named it Mixed-augmentation. We can observe that the Mixed-augmentation method exhibits performance marginally lower than that of the best \textbf{Increase Complexity} approach. However, it can be seen from Table \ref{enlargesize} that with larger data size, mixed-augmentation can reach the best performance. Therefore, it is necessary to mix different types of queries to improve the diversity of augmented questions.  There are more discussions in Appendix \ref{app:augmented method} about improving existing data augmentation methods.


\paragraph{Query and Response Sources}
Here we examine how the query and response quality influence the augmented model performance. We list results in Table \ref{sources}, and draw the following conclusions:
(a) Comparing $\mathcal{\hat{D}}_2^1$ and $\mathcal{D}_1^6$, 
 we find that the queries generated by GPT-4 and GPT-3.5 have no significant impact on SFT performance.
(b) Comparing $\mathcal{D}_1^1$ and $\mathcal{D}_1^6$, we can conclude that when using GPT-4 to generate responses, the temperature has no significant impact on SFT performance.
(c) Comparing $\mathcal{D}_1^1$ and $\mathcal{D}_1^8$, we can conclude that, compared to the zero-shots generation method, the response augmentation prompt we propose plays a substantial role in enhancing the quality of the generated data (+3.6 for LLaMA-7B, +3.7 for LLaMA-2-7B, +3.3 for LLaMA-2-13B). The main reason we consider this is our 1-shot setting stabilizes the response format.
(d) Comparing $\mathcal{D}_1^7$ (25K) and $\mathcal{D}_1^1 \times 0.8$ (24K), we can conclude that, compared to GPT-3.5, the response augmented using GPT-4 yields significantly better results for SFT.

\begin{table}[t]
    \small
    \centering
    \begin{tabular}{lccc}
  \hline
    Model & 7B & 7B-2 & 13B-2  \\
    \hline
    $\mathcal{D}$ & 35.9 & 41.6 & 50.0\\
    \hline
    
    $+\mathcal{D}_1^1 \times 0.8$ & 51.1 & 56.6 & 63.2 \\
    $+\mathcal{D}_1^1$ & 53.0 & 57.0 & 65.5 \\
    $+\mathcal{D}_1^6$ & 51.6 &58.0  &63.8 \\
    $+\mathcal{D}_1^7$ &41.3  &46.7 &52.8\\
    $+\mathcal{D}_1^8$ &49.4 &53.3 &62.2\\
    $+\mathcal{\hat{D}}_2^1$ &52.3 &57.8 &63.3\\
  \hline
  
    \end{tabular}
    \caption{Performance of subsets of AugGSM8K with different query and response sources. $+\mathcal{D}_1^1$ is an omission of $\mathcal{D}+\mathcal{D}_1^1$, and the same notation is used in other tables in this paper.}
    \label{sources}
\end{table}

\subsection{RQ2. What is the Scaling Relationship between the Amount of Augmented Data and Model Performance}\label{sec:relationship}

\begin{figure}[t]
    \centering
    \includegraphics[width=\linewidth]{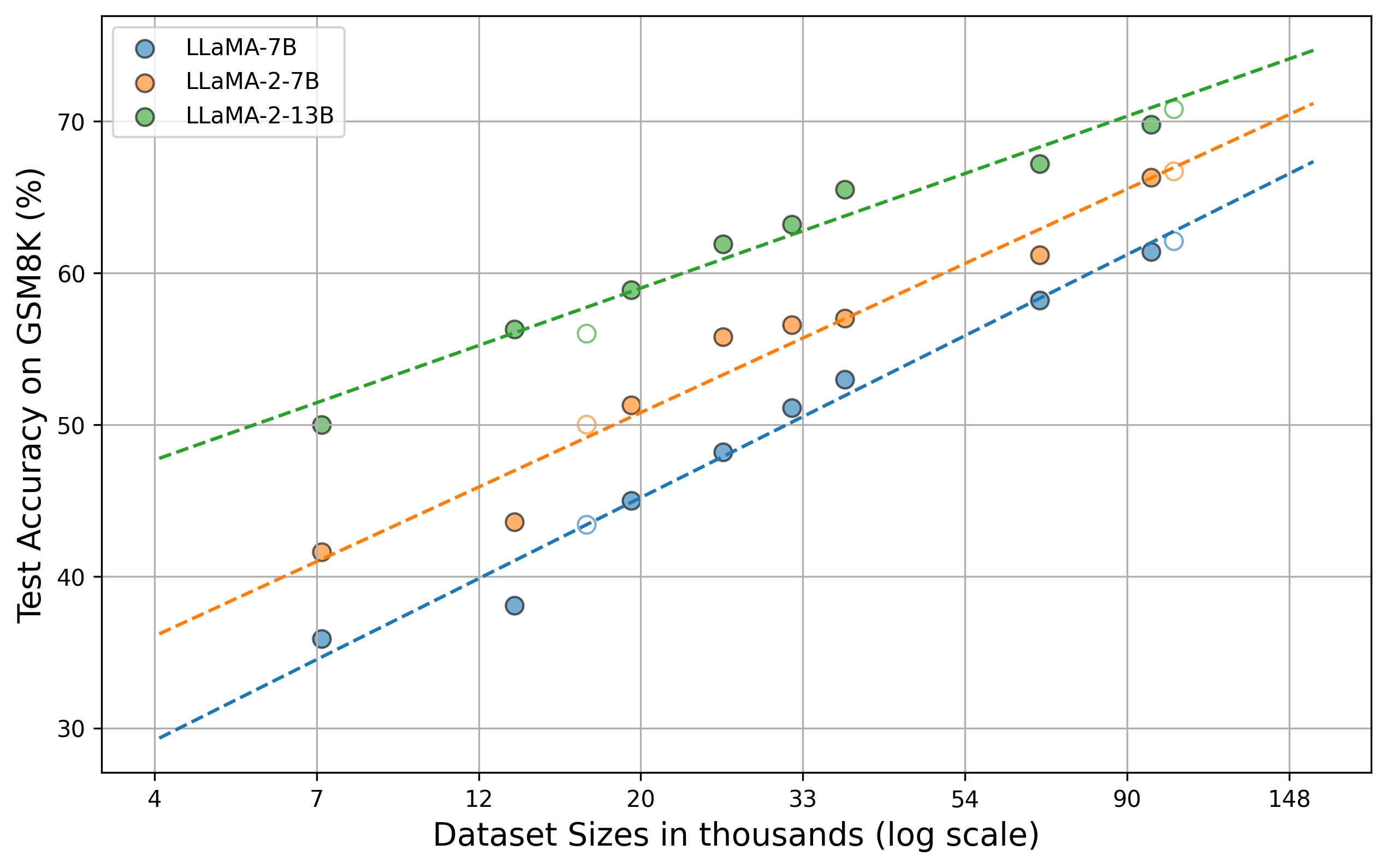}
    \caption{Comparison of test set accuracy on GSM8K for models of varying scales after fine-tuning on AugGSM8K subsets with different query augmentation amount.}
    \label{scaling query}
\end{figure}

\paragraph{Query Augmentation Amount}
We examine how query augmentation amount affects the in-domain performance.
We examine seven data volume configurations including partitioning $\mathcal{D}_1^1$ into proportions of 0, 0.2, 0.4, 0.6, 0.8, 1.0, as well as $\mathcal{D}_1^1 + \mathcal{\hat{D}}_2^1$ and $\mathcal{D}_1^1 + \mathcal{\hat{D}}_2^1 + \mathcal{D}_3^1$ as the augmented datasets. Each augmented query only has one augmented response.
They are mixed with GSM8K $\mathcal{D}$ to apply SFT. 
From Table \ref{scaling query} and Table \ref{tab:query amount}, we can find that within the data volume range of 13-97K, the in-domain performance exhibits a log-linear relationship with the query amount.
We employ linear regression to approximate this relationship. 
As shown in Figure \ref{scaling law}, pre-training models with better initial math reasoning capabilities exhibit a smaller slope which is consistent with \cite{rft}.
This suggests it is harder to improve reasoning ability for a better pre-trained model.
We also conduct validations on our fitted scaling law with an interpolate point at a query amount of 17K ($\mathcal{D} + \mathcal{D}_1^1 \times 0.3$) and an extrapolate point at a query amount of 104K ($\mathcal{D} + \mathcal{D}_1^1 + \mathcal{D}_2^1 + \mathcal{D}_3^1$), discovering that the regression offers accurate predictions of model performances.
We should notice this scaling law cannot be correct within all dataset size ranges since the test set accuracy is bounded.

Besides, the fitted regression shows when \textbf{query augmentation amount} doubles, LLaMA-7B models will improve $10.7 \times \log(2) = 7.4$, LLaMA2-7B will improve $9.8 \times \log(2) = 6.8$, and LLaMA2-13B will improve $7.6 \times \log(2) = 5.3$.
As shown in \citet{rft}, it is estimated that when \textbf{human-written sample amount} doubles, LLaMA-7B models will improve $6.5$ score, LLaMA2-7B will improve $6.6$ score, and LLaMA2-13B models will improve $5.5$ score. 
Query augmentation is similarly effective to human-written samples in term of in-domain performance. 
This demonstrates that query augmentation benefits from the performing proprietary LLMs on GSM8K, thus the sample quality generated by query augmentation is as high as those of human-written samples. When it comes to the relationship of model performance and augmented data size, a segmented log-linear  relationship is presented. The detailed discussion is in Appendix \ref{app:AugMATH}.

\begin{table*}[h]
    \small
    \centering
    \begin{tabular}{lccc}
   \toprule
    Model & 7B & 7B-2 & 13B-2  \\
    \midrule
    Estimation &$y = 10.7\log(x) +13.2$  & $y = 9.8\log(x) 
    +21.3$ & $y = 7.6\log(x) +36.3$\\
    \midrule
    x = 17 prediction &43.4 &49.2 &57.7\\
    x = 17 observation &43.4 &50.0 &56.0 \\
    x = 104 prediction &62.7 &67.0 &71.4\\
    x = 104 observation &62.1 &66.7 &70.8 \\
  \bottomrule
    \end{tabular}
    \caption{The scaling law on amounts of augmented query in GSM8K.}
    \label{scaling law}
\end{table*}

\begin{figure}[h]
    \centering
    \includegraphics[width=\linewidth]{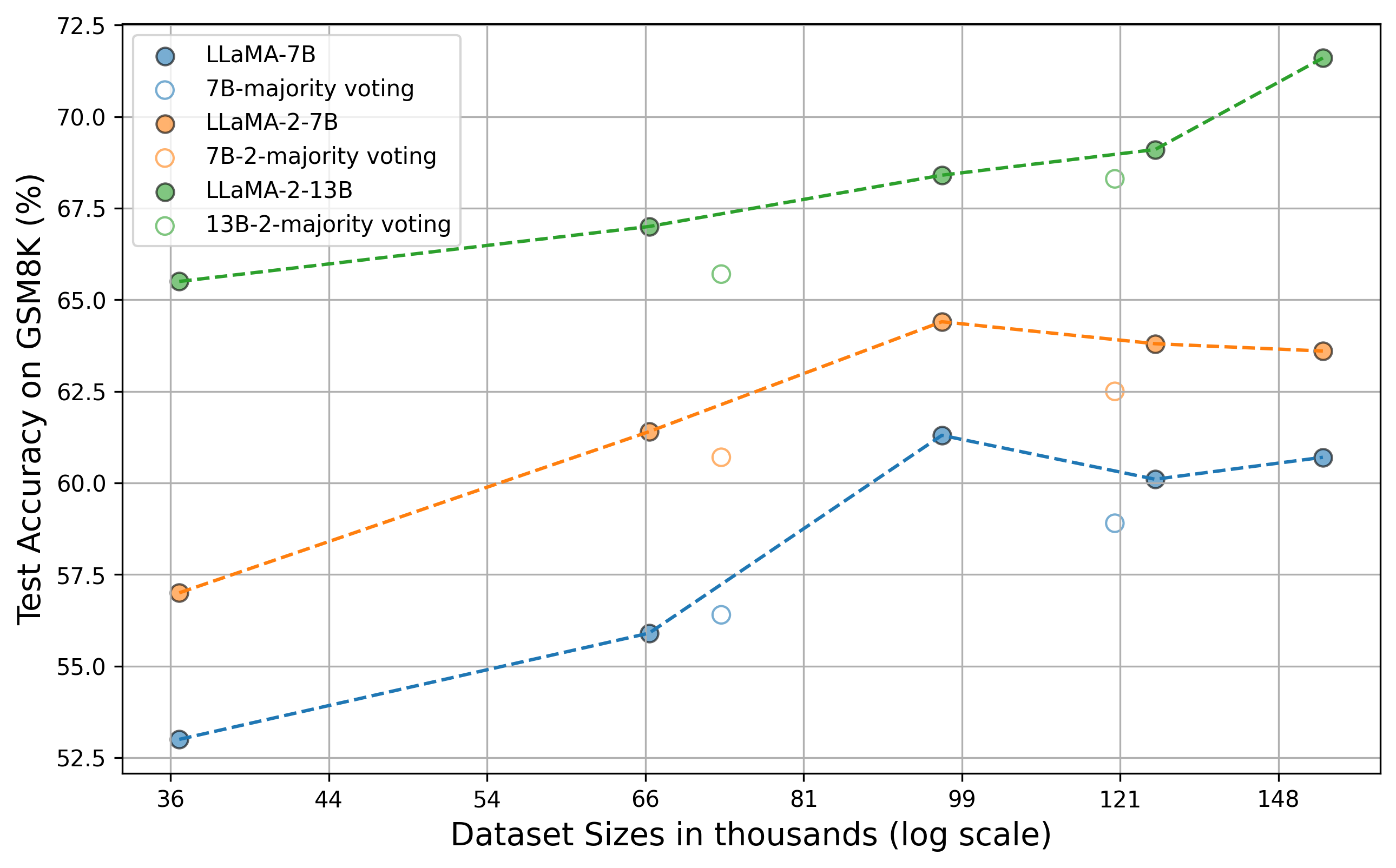}
    \caption{Comparison of test set accuracy on GSM8K for models of varying scales after fine-tuning on AugGSM8K subsets with different response augmentation amount.}
    \label{response amount}
\end{figure}

\paragraph{Response Augmentation Amount}
We further investigate under the data augmentation setting, if we keep the number of queries constant and increase the number of responses, how the in-domain performance changes. We use $\mathcal{D}_1$ as augmented queries and vary the response amount from 1 to 5 per augmented query.
We also try majority voting~\citep{wang2023selfconsistency,huang2022large} to filter the augmented response since we cannot know the correct answer.
In Figure \ref{response amount} and Table \ref{tab:response amount}, we find for LLaMA-7B and LLaMA-2-7B models, once the response data volume reaches 97K (3 responses per query), further increase the number of responses do not yield performance improvement. 
Before this point, model performance improves as the response amount increases.
Thus, query augmentation with accurate responses seems more effective than only response augmentation with the augmented data size scales up.
As for the LLaMA-2-13B model, within the data volume range of 37K to 157K, model performance consistently rises in a roughly linear fashion with increasing data volume, with a slower rate than the 7B model within the ascending interval. 
We then investigate the performance impact brought by majority voting filtering.
If all responses have different answers, we discard the corresponding query-response. Surprisingly, we find that after applying majority voting, the model performance at the same data scale is generally lower than not applying filtering. 
A possible explanation is that even wrong responses generated by GPT-4 are useful for worse models (LLaMA) to improve their abilities. Another explanation is the reduction in the number of queries, as we discard the corresponding query when all response answers are different. To study the relationship between response quality and model performance, we discuss more in Appendix \ref{app:response quality}.


\paragraph{Combination}
We investigate how the combination of query augmentation and response augmentation in AugGSM8K will affect the model’s performance. 
Results are listed in Table \ref{tab:Combination1}. It demonstrates that query augmentation and response augmentation can complement each other to a certain extent to improve in-domain performance.
We conduct SFT on AugGSM8K and AugMATH to obtain MuggleMath which effectively improves the in-domain accuracy on GSM8K and MATH test set in Table \ref{tab:Combination1} and Table \ref{tab:Combination2}. 
These models outperform previous state-of-the-art open-sourced models with a very large margin for 7B and 13B models. \textbf{Model comparisons of MuggleMath and  a broader range of state-of-the-art approaches are in Table \ref{More comparsion}.}
Case studies of MuggleMath are listed in Table \ref{testGSM8K}.
 Moreover, our work suggests that high-quality math training data can be synthesized through powerful large language models by relying solely on simple prompt, without the need for complex designs.

\begin{table*}[h!]
    \small
    \centering
    
    \begin{tabular}{l|cccc}
    \toprule
    Model & 7B & 7B-2 & 13B-2 & 70B-2 \\
    \midrule
    $\mathcal{D}$& 35.9 & 41.6 & 50.0 & 63.2 \\
    RFT~\citep{rft} & 49.1 & 51.2 & 55.3 & 64.8 \\
    WizardMath~\citep{luo2023wizardmath} & - & 54.9 & 63.9 & 81.6 \\
    \midrule
    $+\mathcal{D}_1^1+\mathcal{D}_1^2+\mathcal{D}_1^3$& 61.3 &64.4 &68.4&-\\
    $+\mathcal{D}_1^1+\mathcal{\hat{D}}_2^1+\mathcal{D}_3^1$&61.4 &66.3 &69.8&-\\
    $+\sum_{i=1}^{3}{\mathcal{D}_1^i}+\mathcal{\hat{D}}_2^1+\mathcal{D}_3^1$  &65.4  &68.4 & 74.0  &82.3 \\
    MuggleMath &  \textbf{66.0} & \textbf{69.8}\textcolor{blue}{(+14.9)} & \textbf{74.3}\textcolor{blue}{(+10.4)} & \textbf{82.7}\textcolor{blue}{(+1.1)} \\
    \bottomrule
    \end{tabular}
    \caption{In-domain performance of MuggleMath(AugGSM8K and AugMATH) on GSM8K.}.
    \label{tab:Combination1}
\end{table*}

\begin{table*}[h!]
    \small
    \centering
    
    \begin{tabular}{l|cccc}
    \toprule
    Model & 7B & 7B-2 & 13B-2 & 70B-2 \\
    \midrule
    $\mathcal{D}$& 4.8 & 5.8 & 6.0 & 14.4 \\
    WizardMath~\citep{luo2023wizardmath} & - & 10.7 & 14.0 & 22.7 \\
    MuggleMath &  \textbf{18.4} & \textbf{25.8}\textcolor{blue}{(+15.1)} & \textbf{30.7}\textcolor{blue}{(+16.7)} & \textbf{36.3}\textcolor{blue}{(+13.6)} \\
    \bottomrule
    \end{tabular}
    \caption{In-domain performance of MuggleMath(AugMATH and AugGSM8K) on MATH. See more details of AugMATH in Appendix \ref{app:AugMATH}}.
    \label{tab:Combination2}
\end{table*}

\subsection{RQ3. Can Data Augmentation Incentivize Generalization to Out-of-domain Mathematical Reasoning Tasks?}\label{sec:out-domain}

We have found that query and response augmentation significantly improves in-domain math reasoning performance. 
But we really interested in whether we can improve performances on out-of-domain distribution.
We employ multi-task learning and transfer learning to see how models fine-tuned on AugGSM8K perform on the MATH test set, where we sample 500 questions as test test like~\citep{prm800k}.
We list results in Table \ref{tab:ood}. 
We find that (1) Multi-task learning and transfer learning outperform single-task supervised fine-tuning on LLaMA2-7/13B and do not improve on LLaMA-7B. (2) Although augmenting more query and response can improve GSM8K significantly, it has little to no help in improving MATH performance which indicates that in-domain augmentation data on the GSM8K dataset yields only marginal benefits for the MATH dataset in this setting.
Case studies of models performed on MATH are listed in Table \ref{testMATH}. 

\begin{figure}[t]
    \centering
    \includegraphics[width=\linewidth]{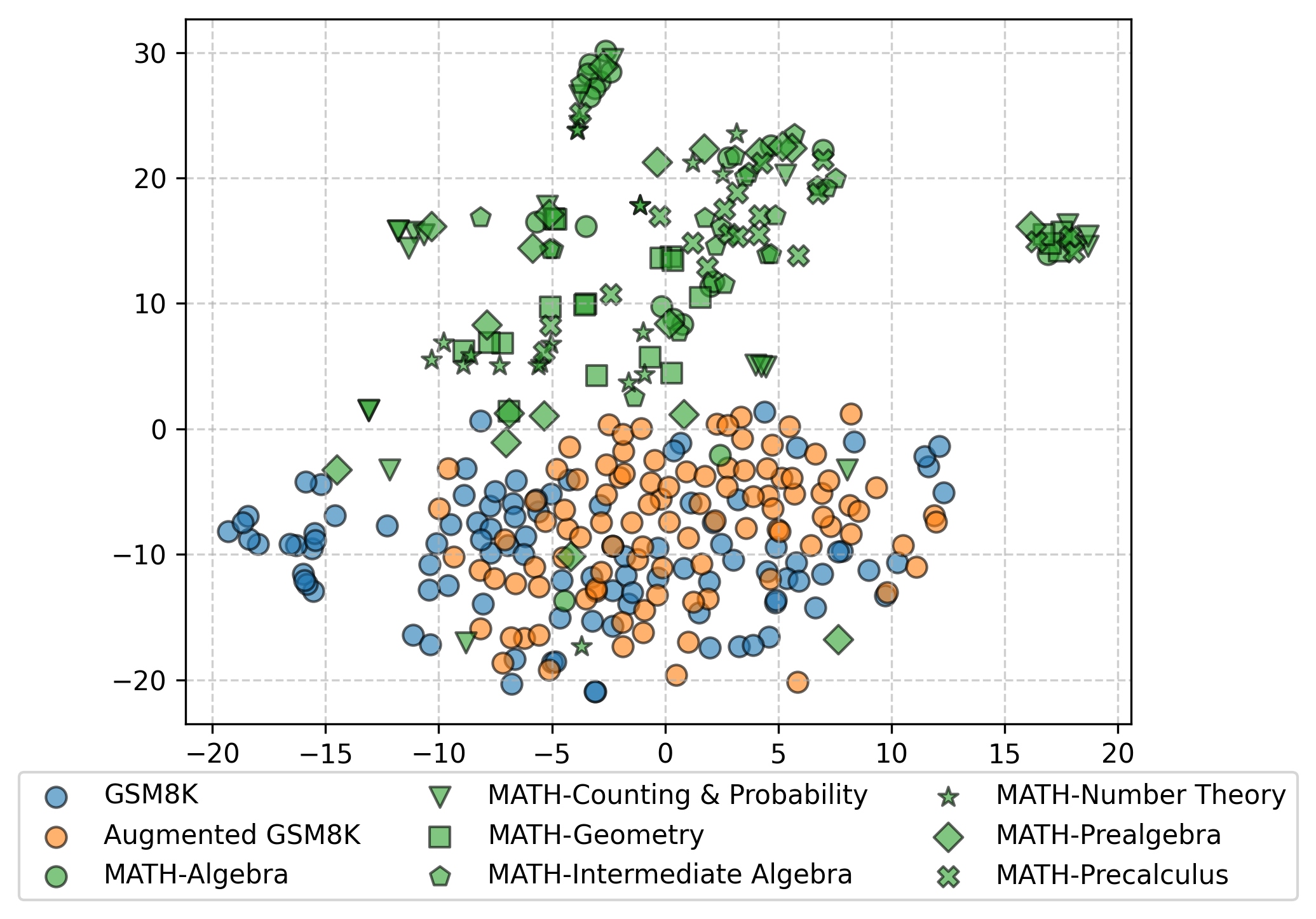}
    \caption{The embedding visualization of queries in GSM8K, MATH and AugGSM8K.}
    \label{query visualization}
\end{figure}

To further investigate why AugGSM8K helps little on the MATH dataset, we use t-SNE in Figure \ref{query visualization} to visualize the hidden representation of problems encoded by LLaMA2-7B, that is the 15-th layer of last token representation of the problem .
We find GSM8K and MATH are separated in hidden space and only some of the problems in MATH are laid in the span of GSM8K.  
The augmented GSM8K problems are laid in the same span of GSM8K which makes sense why it improves little for MATH.

To investiate if the proposed augmentation method improve performance on this subset of MATH, we find that while there is an overlap in the embedding space distribution between MATH and GSM8K, it is relatively small compared with that between GSM8K and AugGSM8K. In the transfer learning setting, training first on GSM8K and then on MATH with LLaMA-13B-2 does provide some benefits for certain subsets, such as Prealgebra, Algebra, and Geometry. However, if we train first on AugGSM8K and then on MATH, this benefit is not only marginal but may even lead to a decrease in performance on other subsets, like Geometry and Prealgebra, which could be related to the data proportions. Overall, the performance improvement on the MATH dataset from augmentation on GSM8K is minimal, even on subsets like Prealgebra, where there is some overlap. For 7B size and multi-task learning setting, we can draw the similar conclusion. Detailed resulsts are listed in Table \ref{subset1} to Table \ref{subset6}.
This suggests if we want to improve math reasoning benchmark performances for LLMs, we can choose to apply augmentation on diverse math subjects. For the generalization from AugMATH to GSM8K, the detailed experiments and explanation are in Appendix \ref{app:AugMATH}. Moreover, we conduct experiments on more out-of-distribution mathematical dataset in Appendix \ref{oodotherdata}. 

\begin{table*}[t]
    \small
    \centering
    \begin{tabular}{lcccc}
    \toprule
    Training Setting & 7B & 7B-2 & 13B-2\\
    \midrule
    \textit{In Context Learning} on MATH  &2.9 &2.5 &3.9 \\
    \textit{Supervised Fine-tuning} on MATH &4.8 &5.8 &6.0\\
    \midrule
    \textit{Multi-task learning} \\
    $+$MATH &4.6 &6.2 &7.6\\
    $+\mathcal{D}_1^1$+MATH &4.8 &4.8 &8.4\\
    \midrule
    \textit{Transfer learning} \\
    $\mathcal{D}$ $\rightarrow$MATH &4.4 &6.0 &9.4\\
    $+\mathcal{D}_1^1\rightarrow$MATH &6.2 &5.6 &7.8\\
    $+\sum_{i=1}^{3}{\mathcal{D}_1^i}+\mathcal{\hat{D}}_2^1+\mathcal{D}_3^1$$\rightarrow$MATH &5.6 &8.4 &9.4\\
    $+\sum_{i=1}^{7}{\mathcal{D}_1^i}$-majority voting$+\mathcal{D}_2^1+\mathcal{D}_3^1$ $\rightarrow$MATH &5.6 &6.0 &9.0\\
    \bottomrule
    \end{tabular}
    \caption{Comparison of test set accuracy on MATH. Multi-task learning means that we fine-tune the models on the mixed dataset of AugGSM8K subset and MATH. Transfer learning means that we first fine-tune the models on subsets of AugGSM8K and then fine-tune on MATH.}
    \label{tab:ood}
\end{table*}

\section{Discussion}
\subsection{Training set vs. Test set accuracy}\label{sec:relation}

Query and response augmentation generate similar problems for the training set which leads to better training set accuracy.
We have shown augmentations improve the accuracy of the in-domain test set.
We want to investigate the relationship between the accuracy of the training set and the test set to find if the accuracy of the training set can be a performance indicator.
We sample 500 samples from the original training set to calculate the accuracy. 
From Figure \ref{train_vs_test}, the training and test accuracy generally exhibit a positive correlation across different augmented data which shows the training accuracy could be an indicator of in-domain performance unless deliberately overfitting.  

\subsection{Make more augmentation on harder problems}\label{sec:hard}

 \begin{figure}[h!]
    \centering
    \vspace{0.3cm}
    \begin{minipage}{0.48\textwidth}
        \centering
        \includegraphics[width=\linewidth]{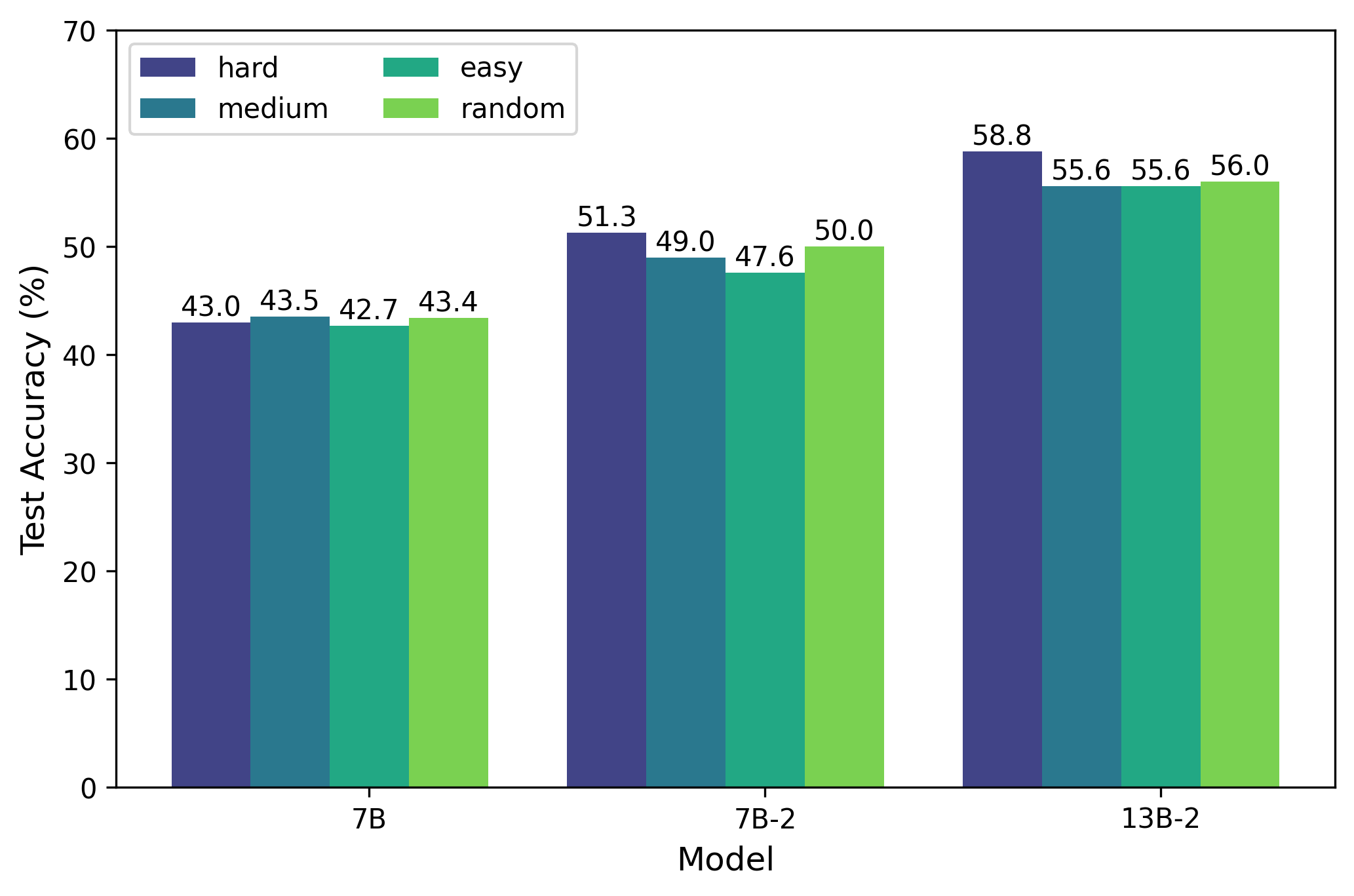}
        \caption{The performance of SFT models with different difficulty augmentation on GSM8K.}
        \label{fig:harder}
    \end{minipage}\hfill
    \begin{minipage}{0.48\textwidth}
        \centering
        \includegraphics[width=\linewidth]{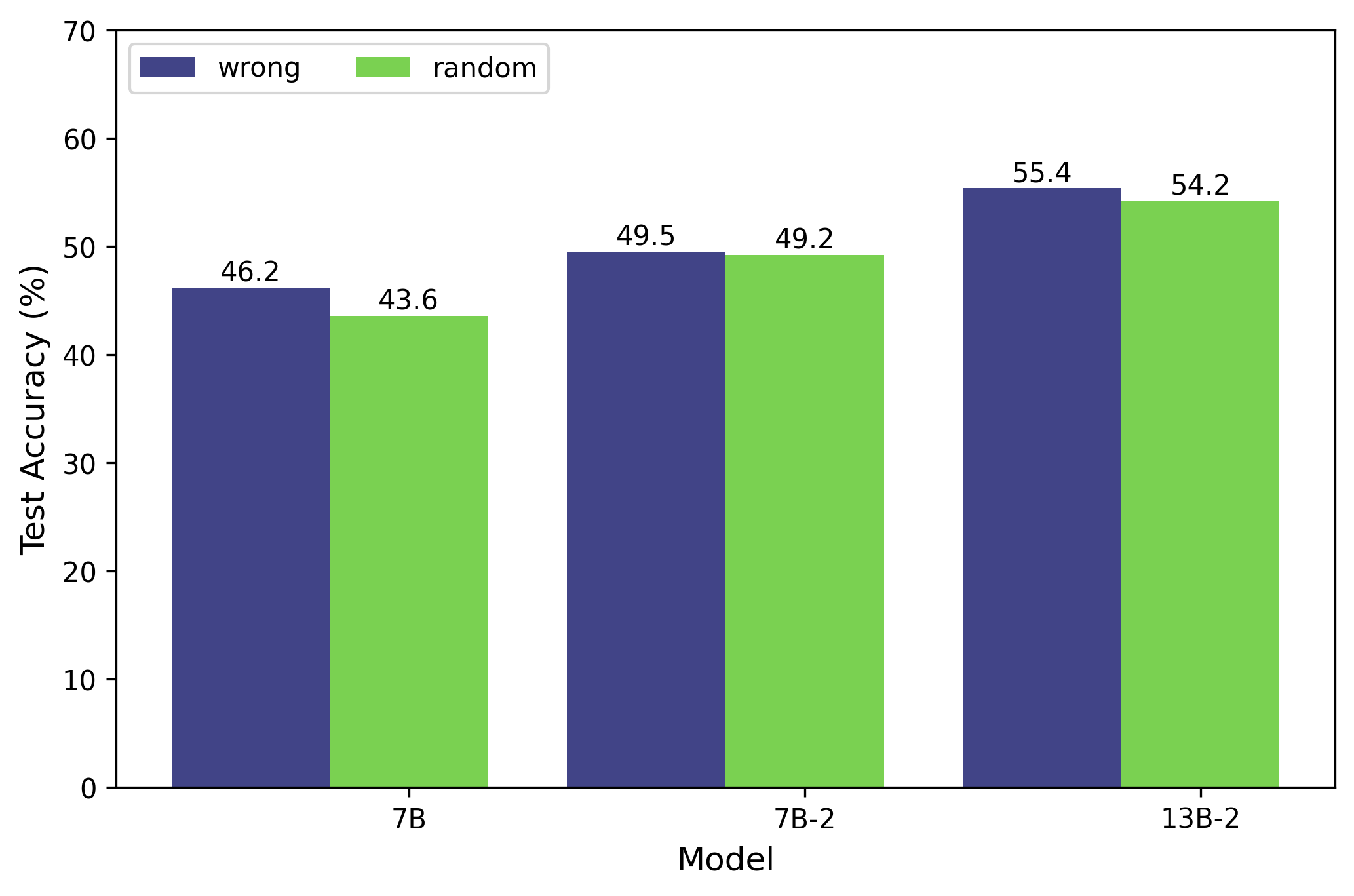}
        \caption{The performance of SFT models with wrong problem augmentation on GSM8K.}
        \label{fig:wrong}
    \end{minipage}
    \vspace{0.6cm}
\end{figure}

During the query augmentation process, it is crucial to understand which kind of queries should be augmented. 
Augmenting too many easy problems may not be effective since the model may have mastered this level of problems.
Here we examine if the model improves more when we augment more on harder or wrong problems.
We define \textbf{hard} problems based on the number of equations, specifically, problems with fewer than three reasoning steps as easy, those with exactly three steps as medium, and those with more than three steps as hard(see more details in Appendix \ref{app:hardproblem}).
We define \textbf{wrong} problems if the SFT model solves them incorrectly.
We apply SFT on subsets of AugGSM8K with augmented queries on easy, medium, hard, wrong, and random problems.
From Figure \ref{fig:harder} and Figure \ref{tab:harder}, it is evident that for LLaMA-2-7B and LLaMA-2-13B, the performance gain from augmenting hard problems is significantly higher than that from augmenting other types of problems.
From Figure \ref{fig:wrong} and Figure \ref{tab:wrong}, we find that augmenting incorrect problems on three models consistently improves more than random query augmentation.
In addition, we have conducted an analysis of our model’s performance on test set problems of varying difficulty. Our analysis of the 7B-2 model’s performance on the test set, shows accuracy rates of 0.55, 0.42, and 0.21 for easy, medium, and hard problems, respectively, while MuggleMath achieves higher accuracy rates of 0.73, 0.70, and 0.64 for the same problem categories. This significant performance boost on difficult questions can be attributed to the fact that the augmented problems we generated are generally more complex than the original problems.

\subsection{Result on the Perturbed Test Set}
We have perturbed two new test sets based on the original GSM8K test set, Change-Test and Aug-Test in in Table \ref{perturb_test}.

Upon evaluating our model on these two perturbed test sets, we found that the performance of MuggleMath consistently and significantly exceeds that of the model fine-tuned on GSM8K alone. This observation suggests that our data augmentation techniques not only enhance the model’s ability to solve the original problems but also contribute to its improved performance on varied and perturbed inputs, thereby indicating a robust generalization capability in in-domain dataset.

\section{Conclusion}

In this paper, we investigate the scaling property of query and response augmentation with respect to math reasoning in-domain and out-of-domain performance.
We find that query and response augmentation can improve in-domain performance very effectively which has a similar improvement to human-written query-response pairs augmentation.
Although we can obtain state-of-the-art in-domain performance by such augmentation, We find that the math reasoning ability improved by the in-domain augmented data is hard to generalized to out-of-domain datasets. Therefore, when using synthetic data, increasing the diversity of the queries to improve the overall mathematical reasoning ability of the LLMs is crucial. In the future, how to use synthetic data to enhance the overall mathematical reasoning ability of the model is a very important research topic.

\section*{Limitations}

In this study, we focused on the domain of mathematical reasoning—a key area of interest for the large language model (LLM) research community. We investigated the effectiveness of data augmentation techniques on both in-domain and out-of-domain  performance. While our work provides insights into the performance scalability and generalizability of Chain-of-Thought (COT) enhanced models in mathematical reasoning, it's important to acknowledge certain limitations: (1)Depth of Generalizability Research: Although our study is among the initial efforts to evaluate the generalizability of COT-augmented models in math reasoning, the study did not extensively explore solutions to enhance out-of-domain performance across a broader spectrum of mathematical reasoning. (2)Unique Data Augmentation Process: We utilized generative models, specifically gpt-3.5-turbo-0613 and gpt-4-0613, for data augmentation, which means that others using the same prompts may not be able to replicate our exact dataset. However, by varying the generation models' temperature setting, we have verified the robustness of our methodology. This suggests that even if the data augmentation process were to be repeated, similar performance enhancements could be achieved, underscoring the reproducibility of the results despite the unique nature of the data generation.
\section{Acknowledgement}
We thank all anonymous reviewers for their helpful comments and suggestions. This research is supported by the National Science and Technology Major Project (2023ZD0121102) and National Natural Science Foundation of China (92270114).

\bibliography{anthology,custom}

\appendix
\newpage
\section{Augmentation on MATH: AugMATH}
\label{app:AugMATH}
\subsection{Query and response augmentation on MATH}
We conducted simple query and response enhancements for MATH whose training set has 7,500 problems. Each query in the MATH dataset was rewritten in five different ways, and we used GPT-4 to sample each variation eight times with a temperature setting of 0.7. Ultimately, we generated $7,500 \times 5 \times
8=300,000$ synthetic MATH data points, Considering that MATH encompasses a broader range of topics—such as elementary calculus and geometry—rather than mainly algebraic problems like GSM8K, we used our GSM8K query augmentation method as a guide and allowed GPT-4 to reference our approach during the query augmentation process without being strictly confined to those five methods. Details on the augmentation prompts can be found in the Appendix \ref{math prompt}. Since the MATH dataset comprises 7,500 questions that include seven subjects—Prealgebra, Algebra, Number Theory, Counting and Probability, Geometry, Intermediate Algebra, and Precalculus—we can decompose MATH into seven subsets each covering different mathematical subjects. Therefore, our augmented dataset, AugMATH, has already undergone augmentation across various mathematical topics. The results on LLaMA-2-7B are listed in Table \ref{mathsubjects}.
\begin{table*}[h!]
\centering
\resizebox{0.98\textwidth}{!}{
\begin{tabular}{lccccccc}
\toprule
SFT dataset & Algebra & Counting \& Probability & Geometry & Intermediate Algebra & Number Theory & Prealgebra & Precalculus \\
\midrule
MATH & 6.7 & 7.4 & 4.6 & 5.3 & 5.6 & 10.8 & 2.7 \\
AugMATH & 33.5 & 20.7 & 17.7 & 9.4 & 19.4 & 35.9 & 12.8 \\
\bottomrule
\end{tabular}
}
\caption{Performance of AugMATH Across Various Mathematical Domains}
\label{mathsubjects}
\end{table*}

\subsection{Relationship of performance and augmented data size}
We conduct SFT of LLaMA-2-7B on subsets of AugMATH, the results are evaluated on MATH test set with 5,000 problems.
\begin{table*}[h!]
\centering
\resizebox{0.98\textwidth}{!}{
\begin{tabular}{lccc}
\toprule
AugMATH + MATH size & Accuracy on MATH Test Set (\%) & Transfer learning to GSM8K\\
\midrule
0 & 2.5 & 40.3 \\
7.5k & 6.5 & 42.5\\
15k & 8.0 & 41.8\\
22.5k & 9.0 &41.9\\
30k & 9.8 &43.1\\
37.5k & 10.6 &41.1\\
82.5k & 14.4 &43.2\\
157.5k & 18.7 &42.8\\
232.5k & 20.3 &46.5\\
307.5k & 23.1 &45.4\\
\bottomrule
\end{tabular}
}
\caption{Performance of different AugMATH + MATH size on the MATH Test Set}
\label{tab:Mugscaling}
\end{table*}

We observe from Table \ref{tab:Mugscaling} a segmented log-linear relationship in the data size's impact on the model's performance. This relationship is evident when the data volume is either less than 37.5k or greater than 82.5k. We conduct log-linear fittings for these two distinct segments separately, denoting $x$ as the data size in thousands and $y$ as the accuracy on the MATH test set. In the range of 7.5k to 37.5k, the log-linear relationship can be fitted as $ y = 1.47 + 2.45\log(x)$, with coefficient of determination 0.992. In the range of 82.5k to 307.5k, it can be fitted as $y = -13.56 + 6.33\log(x)$, with coefficient of determination 0.985.  We can see that once the data volume reaches a certain threshold, the performance gains from doubling the data become much more substantial. 


\subsection{Generalizability of AugMATH to GSM8K}
\label{oodotherdata}
From Table \ref{tab:Mugscaling}, we observe that as the volume of AugMATH data increases, the performance on GSM8K does not continuously rise but rather exhibits significant fluctuations. However, on the whole, when the full AugMATH dataset (300K) is utilized, the fine-tuned performance on GSM8K improves from 40.3 to 45.4. This increase is relatively modest, especially considering that just 30K of AugGSM8K can elevate the model's performance on GSM8K to a score of 58.

In summary, we can see that our proposed simple data augmentation method is equally effective on the MATH dataset, but exhibits a piecewise logarithmic-linear property; although it appears that augmentation on MATH provides certain gains for GSM8K, these gains are very slight relative to the amount of data augmented.

\subsection{Generalizability on other OOD datasets}
To further illustrate generalizability, we conducted tests on five additional datasets: GSM-Hard~\citep{pal}, SVAMP~\citep{svamp}, TabMWP~\citep{tabmvp}, ASDiv~\citep{asdiv}, and MAWPS~\citep{koncel-kedziorski-etal-2016-mawps}.
\begin{table*}[h!]
\centering
\begin{tabular}{lccccc}
\toprule
Model & GSM-Hard & SVAMP & TabMWP & ASDiv & MAWPS \\
\midrule
LLaMA-2-7B & 7.8 & 38.0 & 31.1 & 50.7 & 60.9 \\
LLaMA-2-7B-SFT & 16.1 & 31.9 & 27.8 & 47.4 & 60.0 \\
WizardMath & 20.6 & 57.3 & 38.1 & 59.1 & 73.7 \\
MuggleMATH (AugGSM8K) & 36.5 & 70.9 & 35.6 & 69.5 & 89.8 \\
MuggleMATH (AugMATH) & 17.1 & 53.0 & 41.9 & 59.7 & 68.5 \\
\bottomrule
\end{tabular}
\caption{Model performance on different OOD mathematical datasets}
\label{OODresult}
\end{table*}

It can be seen from Table \ref{OODresult} that MuggleMATH not only outperforms the base model and SFT model on other datasets, but also exhibits a significant performance advantage compared to the state-of-the-art augmentation method, WizardMath. Overall, by performing augmentation on both GSM8K and MATH, we can obtain robust capabilities across a wide range of datasets. The reason behind may be that these out-of-ditribution datasets are actually similar to the distribution of GSM8K and are not so out-of-ditribution.

\section{Instruction prompt for training and inference}
\label{app:alpaca}
Here is the instruction prompt used for the training and inference stage.

\begin{tcolorbox}[
colback=white!10!white,
colframe=black!75!black,
title=Fine-tuning system prompt,
breakable]
Below is an instruction that describes a task. Write a response that appropriately completes the request.\#\#\# Instruction:
 \textbf{**Query.**} \#\#\# Response:
\end{tcolorbox}

\section{Query augmentation prompt for GSM8K}
\label{app:prompt}

Here is the query augmentation prompt we use for GSM8K. We require the models to generate five different augmented problems with our provided example.
We use \texttt{gpt-3.5-turbo-0613} and \texttt{gpt-4-0613} APIs with a temperature of $1.0$ to obtain augmented problems.

\begin{tcolorbox}[
colback=white!10!white,
colframe=black!75!black,
title=Query augmentation prompt,
breakable]
I want you to act as a math teacher. I will provide a grade school math question and you will help to to create more challenging math questions by given ways. Given the question: “James writes a 3-page letter to 2 different friends twice a week. How many pages does he write a year?”, you will modify it by following ideas:\\
 1. \textbf{Change specific numbers}: James writes a 2-page letter to 2 different friends 3 times a week. How many pages does he write in 4 years?\\
 2. \textbf{Introduce fractions or percentages}: James writes a 3-page letter to 2 different friends twice a week. Each week, he adds 50\% more pages to each letter. How many pages does he write in a month?\\
 3. \textbf{Combine multiple concepts}: James writes a 3-page letter to 2 different friends twice a week. He uses both sides of the paper and each side can hold 250 words. If James writes 100 words per minute, how long does it take for him to write all the letters in a week?\\
 4. \textbf{Include a conditional statement}: James writes a 3-page letter to 2 different friends twice a week. If it's a holiday, he writes an additional 5-page letter to each friend. Considering there are 10 holidays in a year, how many pages does he write in a year?\\
 5. \textbf{Increase the complexity of the problem}: James writes a 3-page letter to 2 different friends twice a week. In addition, he writes a 5-page letter to 3 other friends once a week. How many pages does he write in a month, assuming there are 4 weeks in a month?\\
 Now you are given the question:\\
 \textbf{**A new math problem here.**}
\label{tab:query-aug prompt}
\end{tcolorbox}

\section{Response augmentation prompt for GSM8K}
\label{app:response}
We use this prompt to generate responses to ensure the response format which can be viewed as 1-shot setting. We use \texttt{gpt-3.5-turbo-0613} and \texttt{gpt-4-0613} with temperature $0.0$ or $1.0$. 

\begin{tcolorbox}[
colback=white!10!white,
colframe=black!75!black,
title=Response augmentation prompt,
breakable]
\label{response-aug prompt}
I want you to act as an excellent math solver. You will solve the given math question step by step. You need to reply with a python dictionary in the same format as the given examples. Retain decimals to three decimal places. The formulas in the process need to use the format: 48/2 = \textless\textless48/2=24\textgreater\textgreater24 clips. The end of response needs to be: \#\#\#\# \{answer\}.\\
Examples: \{``query'': ``Natalia sold clips to 48 of her friends in April, and then she sold half as many clips in May. How many clips did Natalia sell altogether in April and May?'', ``response'': ``Natalia sold 48/2 = \textless\textless
48/2=24\textgreater\textgreater24 clips in May.Natalia sold 48+24 = \textless\textless48+24=72\textgreater\textgreater72 clips altogether in April and May.\#\#\#\# 72''\}. \\
The given question: \\
\textbf{**A new math problem here.**}
\end{tcolorbox}

\section{Query and response augmentation prompt for MATH}
We use this prompt to augment MATH dataset and  use \texttt{gpt-4-0613} with temperature $0.0$. 
\begin{tcolorbox}[
colback=white!10!white,
colframe=black!75!black,
title=Response augmentation prompt,
breakable]
\label{math prompt}

I want you to act as a math teacher. You should think of some ways to help students do variation training for challenging competition mathematics problems. For example, for a question-solution pair, Question0: James writes a 3-page letter to 2 different friends twice a week. How many pages does he write in a year? Solution0: He writes each friend $3 \times 2=6$ pages a week. So he writes $6 \times 2=12$ pages every week That means he writes $12 \times 52=\boxed{{624}}$ pages a year \#\#end0”  we can propose 5 types of variation exercises, and response with:

    1. Change specific numbers: Question1: James writes a 5-page letter to 3 different friends 4 times a week. How many pages does he write in 3 years? Solution1: To calculate the total number of pages James writes in 3 years, let's first figure out how many pages he writes each week and then multiply that by the number of weeks in 3 years. He writes each friend a 5-page letter, so for 3 friends, that's $5 \times 3 = 15$ pages per writing session. He writes 4 times a week, so the weekly total is $15 \times 4 = 60$ pages. There are 52 weeks in a year, so in one year, he writes $60 \times 52 = 3120$ pages. Finally, over the course of 3 years, he writes $3120 \times 3 = \boxed{{9360}}$ pages. \#\#end1

    2. Introduce fractions or percentages: Question2: James writes a 3-page letter to 2 different friends twice a week. Each week, he adds 100\% more pages to each letter. How many pages does he write in a month? Solution2: Let's take this step by step: In the first week, James writes a 3-page letter to 2 friends twice a week, which is $3 \times 2 \times 2 = 12$ pages in total for the first week. n the second week, he writes 100\% more pages, thus doubling the number of pages in each letter. So he writes $6 \times 2 \times 2 = 24$ pages in total for the second week. In the third week, he again writes double the previous week's pages, so $12 \times 2 \times 2 = 48$ pages in total for the third week. In the fourth week, the number of pages doubles again, which results in $24 \times 2 \times 2 = 96$ pages in total for the fourth week. Now, we'll add up the pages from all four weeks to find out how many pages he writes in a month: $12 + 24 + 48 + 96 = 180$ pages. Therefore, in a month (assuming a 4-week month), James writes $\boxed{{180}}$ pages. \#\#end2

    3. Combine multiple concepts: Question3: James writes a 3-page letter to 2 different friends twice a week. He uses both sides of the paper, and each side can hold 250 words. If James writes at a speed of 100 words per minute, how long does it take him to write all the letters in a week? Solution3: To find out how long it takes James to write all the letters in a week, we first calculate how many words he writes in total. Each letter is 3 pages long, and he writes to 2 friends, which is $3 \times 2 = 6$ pages per writing session. Since he writes twice a week, the total number of pages per week is $6 \times 2 = 12$ pages. Considering each page has two sides and each side holds 250 words, the number of words on one page is $250 \times 2 = 500$ words. Therefore, the total number of words James writes in a week is $500 \times 12 = 6000$ words. Given James writes at a speed of 100 words per minute, the time it takes him to write all the letters in a week is calculated by dividing the total number of words by his writing speed: $6000 \text{{ words}} \div 100 \text{{ words/minute}} = 60$ minutes. So, James takes $\boxed{{60 \text{{ minutes}}}}$ to write all the letters in a week. \#\#end3

    4. Include a conditional statement: Question4: XX Solution4: XX \#\#end4

    5. Increase the complexity of the problem: Question5: XX Solution5: XX \#\#end5

    Now, find five suitable variation training methods for the new problem. Be careful not to let existing methods limit your thinking. Instead, propose variation training methods that are specifically tailored to the given problem: 

    Question0:  \textbf{**A new math problem here.**}

    Solution0:  \textbf{**corresponding solution here here.**}

    Please response with the given example format(including Questions and solutions)
\end{tcolorbox}
\section{Response filter}
\label{app:response filter}
We filter out generated responses by following rules.
\begin{itemize}
    \item Delete the responses without an answer.
    \item Delete the responses that are excessively lengthy$(>1500)$.
    \item Remove superfluous characters beyond the reasoning path and the answer.
\end{itemize}

\section{Case Study of GSM8K}
There are examples of different methods for generating new queries in Table \ref{Multi-query} and different reasoning paths for the same query in Table \ref{Multi-response}. Some examples of MuggleMath-13B answering questions from the GSM8K test set are in Table \ref{testGSM8K} and wrong reasoning processes are labeled in red.

\begin{table*}[h]
\centering
\begin{tabular}{|c|p{8cm}|}
\hline
\textbf{Method} & \textbf{Query} \\
\hline
Original & Weng earns \$12 an hour for babysitting. Yesterday, she just did 50 minutes of babysitting. How much did she earn? \\
\hline
Change specific numbers & Weng earns \$15 an hour for babysitting. Last week, she babysat for 3 hours and 45 minutes every day. How much did she earn in total last week? \\
\hline
Introduce fractions or percentages & Weng earns \$12 an hour for babysitting. However, for every hour she babysits beyond the first 3 hours, she earns an additional 20\% per hour. If she babysat for 5 hours yesterday, how much did she earn? \\
\hline
Combine multiple concepts & Weng earns \$12 an hour for babysitting, but gets paid only 75\% of the total at the time of service. The rest is paid at the end of the month. If she babysat for 3 hours yesterday, how much was she paid immediately and how much will she receive later? \\
\hline
Include a conditional statement & Weng earns \$12 an hour for babysitting. If it's a weekend, she gets a bonus of \$5. If she babysat for 2 hours last Sunday, how much did she earn? \\
\hline
Increase the complexity of the problem & Weng earns \$12 an hour for babysitting during weekdays and \$15 an hour on weekends. If she babysat for 2 hours on Tuesday and 3 hours on Saturday, how much did she earn in total? \\
\hline
\end{tabular}
\caption{Examples of different methods for generating new queries.}
\label{Multi-query}
\end{table*}

\begin{table*}[h]
\centering
\begin{tabular}{|p{3cm}|p{8cm}|}
\hline
Query & Weng earns \$12 an hour for regular babysitting but doubles her rate during weekends. Yesterday, she did 50 minutes of babysitting and it was a weekend. How much did she earn?  \\
\hline
 Response1&Weng earns \$12 per hour and doubles her rate during weekends, so she earned \$12 * 2 = 24 per hour yesterday. There are 60 minutes in an hour, so Weng did 50 minutes / 60 minutes/hour = 0.833 hours of babysitting. Therefore, Weng earned \$24 * 0.833 = 19.992, which we can round to \$20.\\
\hline
Response2& Weng earns \$12*2 = 24 an hour during weekends. 50 minutes is 50/60 = 0.833 of an hour. So, Weng earned 24*0.833 = 19.992 dollars yesterday. \\
\hline
Response3& During weekends, Weng earns 12*2 = 24 dollars per hour for babysitting. There are 60 minutes in an hour. So, 50 minutes is 50/60 = 0.833 of an hour. Weng earned 24*0.833 = 20 dollars for 50 minutes of babysitting during the weekend. \\
\hline
Response4& Her weekend rate is 12*2 = 24 dollars per hour. There are 60 minutes in an hour, so she worked 50/60 = 0.833 hours. So, Weng earned 24*0.833 = 19.992 dollars. \\
\hline
Response5& Weng's rate during weekends is 12*2 = 24 dollars per hour. Since an hour has 60 minutes, Weng earned 24*(50/60) = 20 dollars yesterday. \\
\hline
Response6& Weng earns \$12*2 = 24 an hour during weekends. She worked for 50/60 = 0.833 hour yesterday. So, Weng earned 0.833*\$24 = 19.992 yesterday. \\
\hline
Response7& Weng earns \$12 an hour, but during the weekend, she doubles her rate which means she earns 12*2 = 24 dollars an hour during the weekend. Babysitting for 50 minutes is equivalent to 50/60 = 0.833 hours. Therefore, Weng earned 24*0.833 = 19.992 dollars from babysitting yesterday. \\
\hline
\end{tabular}%

\caption{Different reasoning paths for the same query.}
\label{Multi-response}
\end{table*}

\begin{table*}[h]
\centering
\begin{tabular}{|p{6cm}|p{6cm}|}
\hline
\textbf{Query} & \textbf{Response} \\
\hline
Janet's ducks lay 16 eggs per day. She eats three for breakfast every morning and bakes muffins for her friends every day with four. She sells the remainder at the farmers' market daily for \$2 per fresh duck egg. How much in dollars does she make every day at the farmers' market? & Janet eats 3 eggs for breakfast and bakes 4 eggs into muffins, so she uses 3+4 = 7 eggs every day. Janet's ducks lay 16 eggs per day, so after she uses 7 eggs, she has 16-7 = 9 eggs left. She sells these 9 eggs at the farmers' market for \$2 per egg, so she makes 9*2 = 18 dollars every day at the farmers' market. \\
\hline
A robe takes 2 bolts of blue fiber and half that much white fiber. How many bolts in total does it take? & The white fiber is half the amount of the blue fiber so it takes 2/2 = 1 bolt of white fiber. All total it takes 2+1=3 bolts of fiber. \\
\hline
Josh decides to try flipping a house. He buys a house for \$80,000 and then puts in \$50,000 in repairs. This increased the value of the house by 150\%. How much profit did he make? & He bought the house for 80,000 and put 50,000 in repairs so he had 80,000+50,000 = \$130,000 invested. The value of the house increased by 150\% so it went up \textcolor{red}{130,000*1.5 = \$195,000} in value. He had 130,000 invested and it increased in value by 195,000 so he made a profit of 195,000-130,000 = \$65,000. \\
\hline
James decides to run 3 sprints 3 times a week. He runs 60 meters each sprint. How many total meters does he run a week? & He runs 3*60=180 meters per day. So he runs 180*3=540 meters. \\
\hline
Every day, Wendi feeds each of her chickens three cups of mixed chicken feed, containing seeds, mealworms and vegetables to help keep them healthy. She gives the chickens their feed in three separate meals. In the morning, she gives her flock of chickens 15 cups of feed. In the afternoon, she gives her chickens another 25 cups of feed. How many cups of feed does she need to give her chickens in the final meal of the day if the size of Wendi's flock is 20 chickens? & Every day, Wendi feeds her chickens a total of \textcolor{red}{15+25=40} cups of feed. Thus, she must feed her chickens 40/3=13.333 cups of feed in the final meal of the day. \\
\hline
\end{tabular}
\caption{Some examples of MuggleMath-13B answering questions from the GSM8K test set. Wrong reasoning processes are labeled in red.}
\label{testGSM8K}
\end{table*}

\begin{table*}[h]
\centering
\begin{tabular}{|p{6cm}|p{6cm}|}
\hline
\textbf{Instruction} & \textbf{Response} \\
\hline
A regular hexagon can be divided into six equilateral triangles. If the perimeter of one of the triangles is 21 inches, what is the perimeter, in inches, of the regular hexagon? & Since the triangle is equilateral, each side length is 7 inches.  The hexagon has six sides of equal length, so its perimeter is $6\times 7=\boxed{42}$ inches. \\
\hline
What is the smallest positive perfect cube that can be written as the sum of three consecutive integers? & \textcolor{red}{The response provided is incomplete and does not provide a valid answer to the question.} \\
\hline
The set of points $(x,y,z)$ that satisfy $2x = 3y = -z$ is a line. The set of points $(x,y,z)$ that satisfy $6x = -y = -4z$ is another line. Find the angle between these lines, in degrees. & The direction vector of the first line is $\textcolor{red}{<2,3,-1>}$, and the direction vector of the second line is $\textcolor{red}{<6,-1,-4>}$.  The cosine of the angle between these direction vectors is $\frac{<2,3,-1> \cdot <6,-1,-4>}{\left\| <2,3,-1> \right\| \left\| <6,-1,-4> \right\|} = \textcolor{red}{\frac{12}{20}} = \frac{3}{5}.$ Therefore, the angle between the lines is $\boxed{\textcolor{red}{75^\circ}}.$ \\
\hline
What is the distance, in units, between the points $(2, -6)$ and $(-4, 3)$? Express your answer in simplest radical form. & We use the distance formula: $\sqrt{(2 - (-4))^2 + (-6 - 3)^2} = \sqrt{\textcolor{red}{16} + 9^2} = \sqrt{16 + 81} = \sqrt{97} = \boxed{\textcolor{red}{3\sqrt{7}}}$. \\
\hline
The expression $2\cdot 3 \cdot 4\cdot 5+1$ is equal to 121, since multiplication is carried out before addition. However, we can obtain values other than 121 for this expression if we are allowed to change it by inserting parentheses. For example, we can obtain 144 by writing $(2\cdot (3\cdot 4)) \cdot (5+1) = 144$. In total, how many values can be obtained from the expression $2\cdot 3\cdot 4 \cdot 5 + 1$ by inserting parentheses? (Note that rearranging terms is not allowed, only inserting parentheses). & We can obtain 121 by not inserting any parentheses, so we need to insert at least one pair of parentheses to obtain a different value. We can obtain 144 by inserting one pair of parentheses as shown above. If we insert two pairs of parentheses, we get $(2\cdot (3\cdot (4\cdot 5))) + 1 = 121$. \textcolor{red}{Thus, we can obtain 121 by inserting two pairs of parentheses, so we need to insert at least three pairs of parentheses to obtain a different value. We can obtain 168 by inserting three pairs of parentheses, since $(2\cdot (3\cdot (4\cdot (5+1)))) = 168$.} Thus, there are $\boxed{\textcolor{red}{4}}$ values that can be obtained from the expression $2\cdot 3\cdot 4\cdot 5 + 1$ by inserting parentheses. \\
\hline
\end{tabular}
\caption{Some examples of MuggleMath-13B answering questions from the MATH test set. Wrong reasoning processes are labeled in red.}
\label{testMATH}
\end{table*}

\section{Difficulty level definition on GSM8K}
\label{app:hardproblem}
We conducted a statistical analysis of the reasoning paths required for 7,473 questions in the GSM8K training set, categorizing them as hard, medium, and easy. Specifically, we defined questions with more than three formulas, exactly three formulas, and less than three formulas as hard, medium, and easy, respectively. This categorization yielded a balanced distribution with 2,357 easy, 2,360 medium, and 2,756 hard problems. This approach ensures a relatively equal number of problems in each category.


\section{Discussion of augmentation methods for GSM8K}
\label{app:augmented method}
\subsection{How to improve existing data augmentation methods?}

If one wants to further improve using these ideas, there are two relatively direct approaches to enhance our current methodology.
\begin{itemize}
    \item The first approach involves thinking of additional methods beyond the initial five. Although it is impossible to exhaust all the methods of rewriting queries, it is relatively easy to propose some similar rewriting approaches and examples, akin to the five schemes we have suggested. For example, "Introduce unknown quantities", "Introduce a range", "Finding averages" may be reasonable. Limited by resources, we only employ "Introduce unknown quantities" to expand the dataset. 
    \item The second approach is to augment the queries we have expanded again. We conducted further enhancement on "mixed-augmentation" while maintaining the same data volume. We called it "Mixed-augmentation-second".
\end{itemize}

Taking the prompt in the question as an instance, the original problem is: "James writes a 3-page letter to 2 different friends twice a week. How many pages does he write a year?" Our five methods are as follows:
\begin{itemize}
    \item Change specific numbers: James writes a 2-page letter to 2 different friends 3 times a week. How many pages does he write in 4 years?
    \item Introduce fractions or percentages: James writes a 3-page letter to 2 different friends twice a week. Each week, he adds 50\% more pages to each letter. How many pages does he write in a month?
    \item Combine multiple concepts: James writes a 3-page letter to 2 different friends twice a week. He uses both sides of the paper, and each side can hold 250 words. If James writes 100 words per minute, how long does it take for him to write all the letters in a week?
    \item Include a conditional statement: James writes a 3-page letter to 2 different friends twice a week. If it's a holiday, he writes an additional 5-page letter to each friend. Considering there are 10 holidays in a year, how many pages does he write in a year?
    \item Increase the complexity of the problem: James writes a 3-page letter to 2 different friends twice a week. In addition, he writes a 5-page letter to 3 other friends once a week. How many pages does he write in a month, assuming there are 4 weeks in a month?

    In addition to these, we can conceive of other methods:

    \item Introduce unknown quantities: James writes some pages to his friends every week. After 4 years, he has written 1,440 pages. How many pages does James write each week?

    \item Introduce a range: James writes about 150-200 words per page. If he writes a 4-page letter, what is the maximum and minimum number of total words he writes?

    \item Finding Averages: James wrote 5 letters last week. The number of pages were 2, 3, 4, 3, and 2, respectively. What was the average number of pages per letter?
\end{itemize}

\begin{table*}[h!]
\centering
\begin{tabular}{lccc}
\toprule
Query aug. type & LLaMA-7B & LLaMA-2-7B & LLaMA-2-13B \\
\midrule
Change numbers & 41.5 & 48.5 & 54.1 \\
Fractions or percentages & 41.2 & 46.2 & 54.4 \\
Combine multiple concepts & 41.1 & 47.5 & 56.1 \\
Conditional statement & 41.7 & 45.8 & 56.4 \\
Increase complexity & 42.4 & 48.6 & 57.6 \\
Introduce unknown quantities & 41.5 & 47.2 & 56.3 \\
Mixed-augmentation & 42.2 & 48.2 & 57.2 \\
Mixed-augmentation-second & 40.5 & 46.0 & 53.3 \\
\bottomrule
\end{tabular}
\caption{Comparison of different query augmentation types across models on GSM8K}
\label{tab:second-mix}
\end{table*}

We can observe that the "Introduce Unknown Quantities" strategy is essentially as effective as the other five methods from Table \ref{tab:second-mix}. Continuing to augment the expanded queries, however, yields slightly less effective results compared to the original five methods. A possible reason for this could be that the problems created through secondary augmentation are too complex, resulting in a lower accuracy rate in the responses provided by GPT-4, thereby impacting the efficacy of the augmentation.

\subsection{Will mixed-augmentation has advantages over increase the complexity if we enlarge the size of data?}

Increasing the dataset size to observe whether the mixed-augmentation approach can yield better results due to its diverse enhancements is indeed necessary. We categorized the data of different query augmentations in $\mathcal{D}_1^1$
, $\mathcal{D}_2^1$
,$\mathcal{D}_3^1$ and performed random sampling from these mixed datasets to obtain a comparable quantity of data. We conduct a SFT on LLaMA-2-7B, the results are list in Table \ref{tab:enlargesize}.

\begin{table*}[h!]
\centering
\begin{tabular}{lcc}
\toprule
Query aug. type & Augmented data: 6.3K & Augmented data: 19K \\
\midrule
Change numbers & 48.5 & 53.4 \\
Fractions or percentages & 46.2 & 52.7 \\
Combine multiple concepts & 47.5 & 54.8 \\
Conditional statement & 45.8 & 53.7 \\
Increase complexity & 48.6 & 54.0 \\
Mixed-augmentation & 48.2 & 55.3 \\
\bottomrule
\end{tabular}
\caption{The performance comparision of different augmentation strategies when we enlarge the data size.}
\label{tab:enlargesize}
\end{table*}

\section{ The effect of augmentation response quality of GSM8K}
\label{app:response quality}
\subsection{Under what circumstances can wrong answers have a positive effect?}
To investigate the effect of augmentation on response quality, we aim to collect datasets with responses that are uniformly incorrect, uniformly correct, and partially correct. We amalgamate $\mathcal{D}_1^1$
, $\mathcal{D}_1^2$
,$\mathcal{D}_1^3$,$\mathcal{D}_1^4$
, $\mathcal{D}_1^5$
,$\mathcal{D}_1^6$,$\mathcal{D}_1^7$,
 and categorize the combined dataset into three groups:
 \begin{itemize}
     \item The first category consists of cases where all seven responses are the same.
     \item The second category includes instances where each of the seven responses is distinct.
     \item The third category encompasses queries with 2 to 6 varied answers.
 \end{itemize}

To mitigate the influence of the query, we eliminate the first and second categories and focus on the third category, where we vote on the answers. The answer with the majority of votes is deemed the correct response, while the others are classified as incorrect. We randomly select one correct and one incorrect response for each query from the third category to form the sets $\mathcal{D}_{\textbf{corrct}}$
 and $\mathcal{D}_{\textbf{wrong}}$
, ensuring both sets contain identical queries. Subsequently, we randomly sample half of the queries from $\mathcal{D}_{\textbf{corrct}}$
 and pair the remaining queries from $\mathcal{D}_{\textbf{wrong}}$
 to create $\mathcal{D}_{\textbf{half}}$
, which shares the same queries as 
$\mathcal{D}_{\textbf{corrct}}$
 and $\mathcal{D}_{\textbf{wrong}}$ in Table \ref{tab:estimatedacc}.
 Although the estimated accuracy may not be precise, it serves to indicate the trend.

\begin{table*}[t]
\centering
\begin{tabular}{lccc}
\toprule
Dataset &$\mathcal{D}_{\textbf{corrct}}$  &$\mathcal{D}_{\textbf{wrong}}$  & $\mathcal{D}_{\textbf{half}}$  \\
\midrule
Size & 20.5k & 20.5k & 20.5k \\
Estimated accuracy & 100\% & 0\% & 50\% \\
SFT LLaMA-2-7B-SFT & 58.7 & 46.1 & 52.4 \\
SFT MuggleMATH-7B & 67.0 & 54.7 & 57.0 \\
\bottomrule
\end{tabular}
\caption{Dataset Size and Estimated Accuracy}
\label{tab:estimatedacc}
\end{table*}

We conduct Supervised Fine-Tuning (SFT) on these three datasets on two models with varying accuracy: LLaMA-2-7B-SFT (41.6 on GSM8K test set), which is fine-tuned from the GSM8K training set, and MuggleMATH-7B (68.4 on GSM8K test set), which is fine-tuned from five subsets of AugGSM8K. The results are as follows:


From Table \ref{tab:estimatedacc}, we can draw some conclusions: (1)For LLaMA-2-7B-SFT, the higher the accuracy of the fine-tuned dataset, the more substantial the performance gain. There is an improvement of 4.5 percentage points even when fine-tuned on a dataset consisting solely of incorrect responses. This may be attributable to the presence of correct reasoning steps within the responses with wrong answers, as mentioned in~\citep{prm800k}. (2)For MuggleMATH-7B, the lower the accuracy of the dataset, the more significant the performance degradation.

In summary, erroneous data still contributes to performance improvement for models with poorer performance, but it has a detrimental effect on models with superior performance.
\subsection{The relationship between response quality and data volume.}
To delve deeper into the relationship between the quality and quantity of augmented data, we partition $\mathcal{D}_{\textbf{corrct}}$ ,$\mathcal{D}_{\textbf{wrong}}$ and $\mathcal{D}_{\textbf{half}}$
 into different fractions and conduct Supervised Fine-Tuning (SFT) on LLaMA-2-7B-SFT (41.6 on GSM8K test set).

 \begin{table*}[h!]
\centering
\begin{tabular}{lcccc}
\toprule
 & $\mathcal{D}_{\textbf{wrong}}$ & $\mathcal{D}_{\textbf{wrong}}/2$ & $\mathcal{D}_{\textbf{wrong}}$/4 & $\mathcal{D}_{\textbf{wrong}}$/8 \\
\midrule

LLaMA-2-7B-SFT & 46.1 & 45 & 41.7 & 40.3 \\
\toprule
 & $\mathcal{D}_{\textbf{half}}$ & $\mathcal{D}_{\textbf{half}}/2$ & $\mathcal{D}_{\textbf{half}}$/4 & $\mathcal{D}_{\textbf{half}}$/8 \\
 \toprule
LLaMA-2-7B-SFT & 52.4 & 48.7 & 45.8 & 40.6 \\
\toprule
 & $\mathcal{D}_{\textbf{correct}}$ & $\mathcal{D}_{\textbf{correct}}/2$ & $\mathcal{D}_{\textbf{correct}}$/4 & $\mathcal{D}_{\textbf{correct}}$/8 \\
 \toprule
LLaMA-2-7B-SFT & 58.7 & 53.4 & 47.5 & 42.5 \\
\bottomrule
\end{tabular}
\caption{LLaMA-2-7B-SFT Performance Across Different datasets with various accuracy on GSM8K}
\label{tab:scalingwrong}
\end{table*}

From these results in table \ref{tab:scalingwrong}, we can draw some insightful conclusions: (1)As the volume of data is increased, even datasets of poor quality can still yield performance gains for LLaMA-2-7B-SFT. (2)The higher the quality of the responses, the more substantial the performance improvements achieved with each doubling of the data volume.

\subsection{The Accuracy and difficulty of different query augmentation categories}

In an effort to discern the performance disparities among various query augmentation techniques, we evaluate the accuracy of responses across different categories. By employing majority voting to determine the reference answer from $\mathcal{D}_1^1$, $\mathcal{D}_1^2$, $\mathcal{D}_1^3$, $\mathcal{D}_1^4$, $\mathcal{D}_1^5$, $\mathcal{D}_1^6$ and $\mathcal{D}_1^7$
, we consider queries that yield seven distinct answers as incorrect. Additionally, we compute the average number of reasoning steps for each augmentation method—a formula such as "30+90 = <<30+90=120>>120" is counted as a single step.

\begin{table*}[h!]
\centering
\resizebox{0.98\textwidth}{!}{
\begin{tabular}{lccccc}
\toprule
Query aug. type & LLaMA-7B & LLaMA-2-7B & LLaMA-2-13B & Accuracy (\%) & Average reasoning step \\
\midrule
Change numbers & 41.5 & 48.5 & 54.1 & 87.02 & 3.31 \\
Fractions or percentages & 41.2 & 46.2 & 54.4 & 73.13 & 4.11 \\
Combine multiple concepts & 41.1 & 47.5 & 56.1 & 67.89 & 4.71 \\
Conditional statement & 41.7 & 45.8 & 56.4 & 66.72 & 4.49 \\
Increase complexity & 42.4 & 48.6 & 57.6 & 62.08 & 5.24 \\
Mixed-augmentation & 42.2 & 48.2 & 57.2 & 71.53 & 4.37 \\
\bottomrule
\end{tabular}
}
\caption{Comparison of Different Query Augmentation Types for their performance and difficulty}
\label{tab:reasongsteps}
\end{table*}

From the Table \ref{tab:reasongsteps}, we observe a general trade-off between the number of reasoning steps and accuracy. As corroborated in section 4.2, the augmentation yields greater benefits for more challenging problems. Notably, the "Increase Complexity" method, while exhibiting the lowest dataset accuracy, involves the highest number of reasoning steps. This suggests that augmenting with more complex problems can lead to more substantial benefits for the model.

\section{A more detailed comparison of GSM8K, MATH, AugGSM8K, and AugMATH}

\begin{table*}[h!]
\centering
\resizebox{0.98\textwidth}{!}{
\begin{tabular}{lcccccc}
\toprule
Dataset & Annotation & Samples & Fields & Average Reasoning Steps & Language Style \\
\midrule
GSM8K & Human & 7473 & Pre-Algebra & 3.17 & human-like writing with formula computation in \texttt{<<>>} \\
MATH & Human & 7500 & Pre-Algebra; Inter-Algebra; Algebra; Probability;  NumTheory; Pre-calculus; Geometry & 7.61 & LaTeX style \\
AugGSM8K & GPT-3.5 \& GPT-4 & 330,000 & Pre-Algebra & 4.37 & human-like writing with formula computation in \texttt{<<>>} \\
AugMATH & GPT-4 & 300,000 & Pre-Algebra; Inter-Algebra; Algebra; Probability;  NumTheory; Pre-calculus; Geometry & 4.88 & LaTeX style \\
\bottomrule
\end{tabular}
}
\caption{Datasets Comparision}
\end{table*}

In GSM8K and AugGSM8K, a formula such as "30+90 = <<30+90=120>>120" is counted as a single step.

Finding a reasonable way to calculate reasoning steps is challenging in the MATH and AugMATH datasets, as the answers in the MATH dataset are in LaTeX format and vary greatly. We consider one sentence in the response (ending with a period or semicolon) as one step of reasoning. An interesting finding is that the number of reasoning steps in AugMATH is significantly lower than in MATH. One possible explanation could be an issue of linguistic style; GPT-4 and humans might use a different number of sentences for the same number of reasoning steps. Therefore, the reasoning steps in MATH are difficult to compare. The second reason may be that GPT-4 has only a 42\% accuracy rate on MATH problems, and on the more challenging MuggleMATH, it provides incorrect responses for most questions. These guys incorrect responses may have fewer reasoning steps than are actually required for the problems.

The reason why a model fails to generalize from one dataset to another after data augmentation might be influenced by many factors. From the perspective of human learning, no matter how much one studies elementary school mathematics, it would be quite difficult to solve high school math problems. Therefore, we believe that simply performing data augmentation during the SFT (Supervised Fine-Tuning) phase to increase accuracy on a dataset, while it may yield significant performance gains, might not lead to as substantial improvements for large models across a wide variety of reasoning tasks as it may appear.

\section{Detailed Experimental results}
\label{added_re}
We list the detailed experimental results of different settings here.
\begin{figure}[t]
\centering
\centering
\includegraphics[width=2.7in]{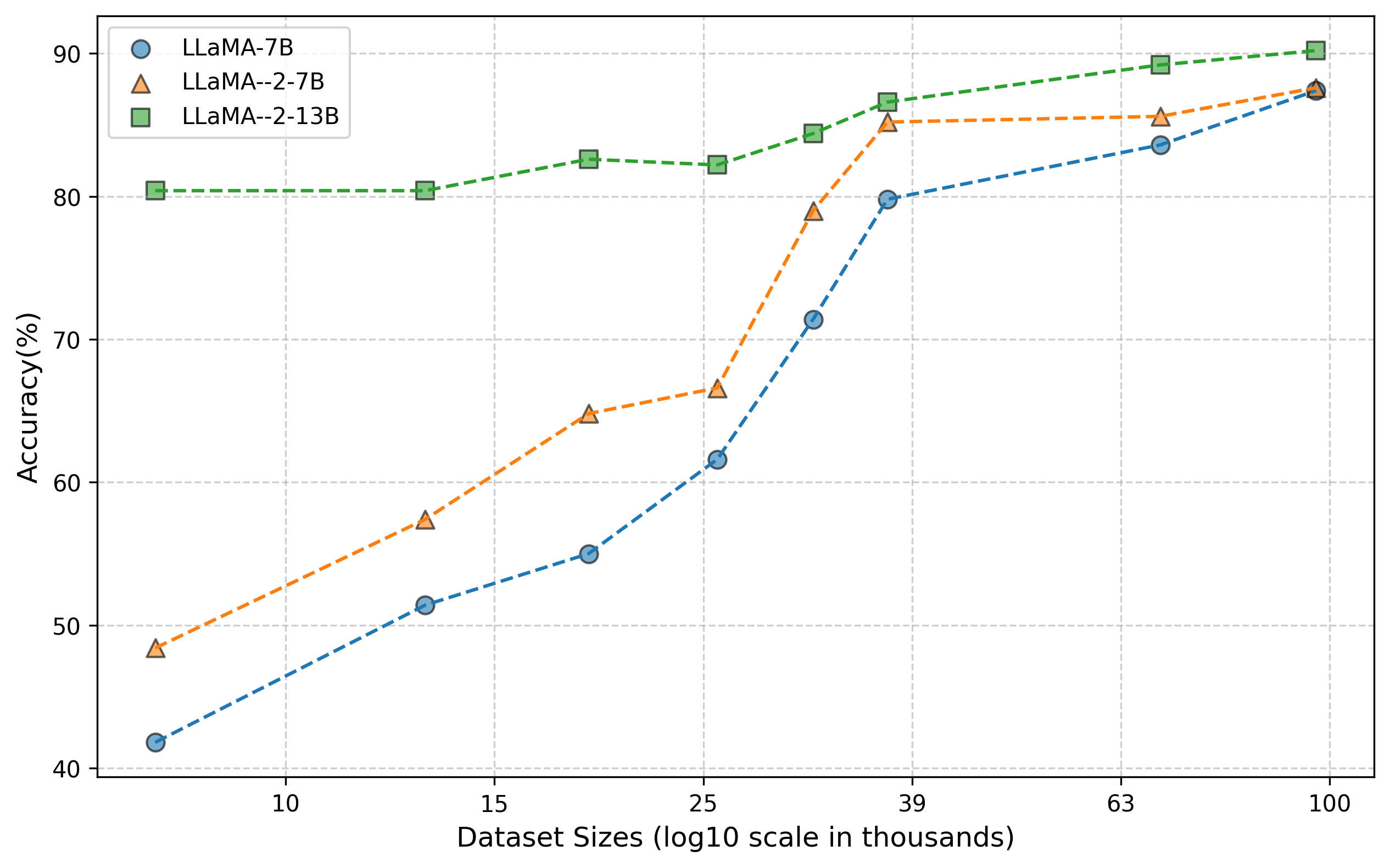}
\caption{The relationship of accuracy on the training set and the amounts of augmentation data.}
\label{scaling train}
\end{figure}
\begin{figure}[h]
\centering
\includegraphics[width=2.7in]{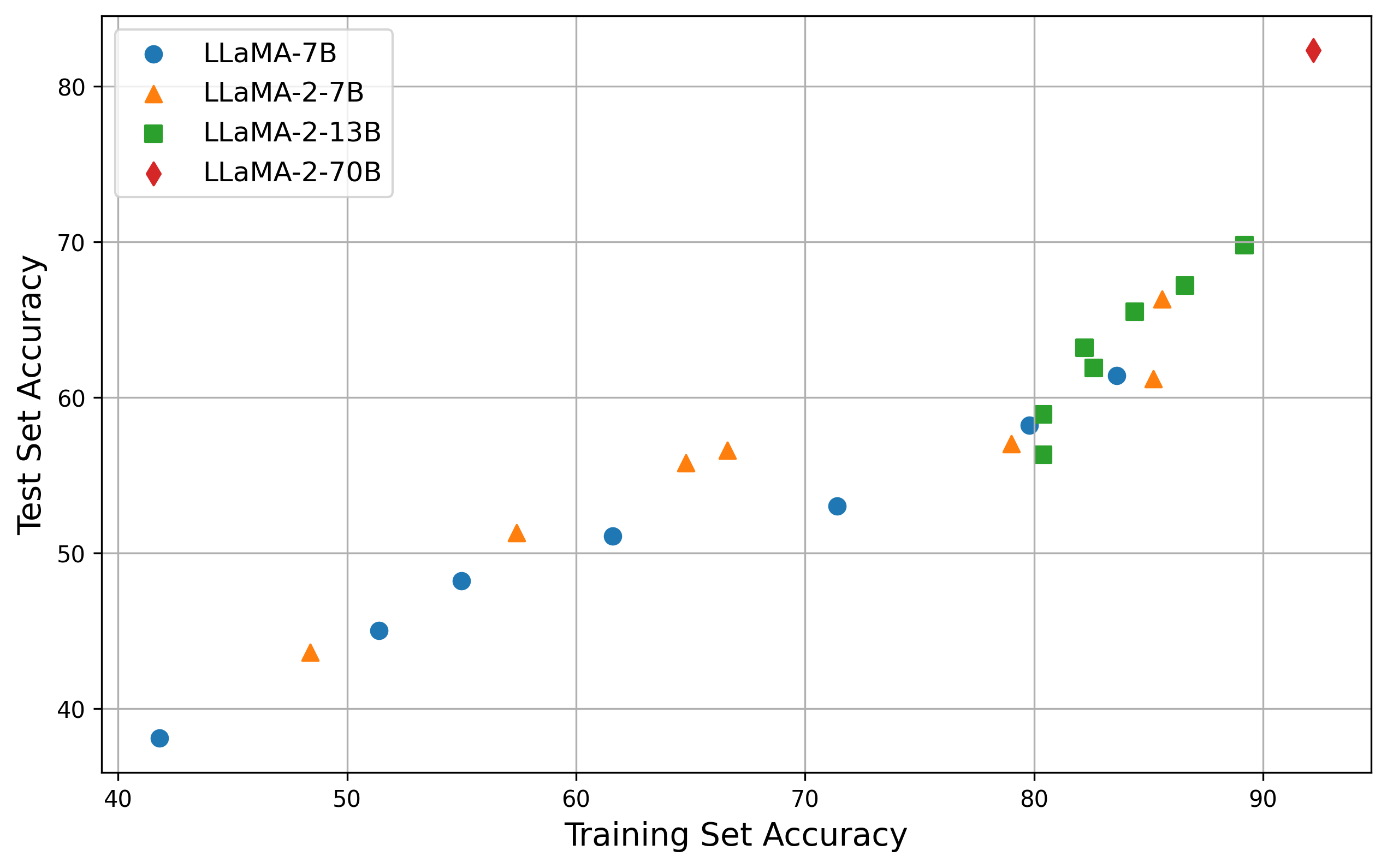}
    \caption{The relationship of accuracy on the GSM8K test set and the original GSM8K training set.}
    \label{train_vs_test}
\end{figure}

\begin{table*}[h]
    \small
    \centering
    \begin{tabular}{lccc}
  \hline
    Query aug. type & 7B & 7B-2 & 13B-2  \\
    \hline
    No aug. ($\mathcal{D}$) & 35.9 & 41.6 & 50.0 \\
    \hline
    Change numbers &41.5 &48.5 &54.1\\
    Fractions or percentages & 41.2 &46.2 &54.4\\
    Combine multiple concepts & 41.1 &47.5 & 56.1\\
    Conditional statement & 41.7 &45.8 &56.4\\
    Increase complexity & 42.4 &48.6 & 57.6\\
    Mixed-augmentation	&42.2	&48.2	&57.2\\
    \hline
    All aug. ($\mathcal{D} + \mathcal{D}_1^1$) & \textbf{53.0} & \textbf{57.0} & \textbf{65.5} \\
  \hline
    \end{tabular}
    \caption{Different query augmentation strategies on GSM8K performances.}
    \label{tab:query type}
\end{table*}

\begin{table*}[h]
    \small
    \centering
    
    \begin{tabular}{lccc}
  \hline
    Query aug. & 7B & 7B-2 & 13B-2  \\
    \hline
    $\mathcal{D}$ & 35.9 & 41.6 & 50.0\\
    \hline
    $+\mathcal{D}_1^1 \times 0.2$ &38.1 &43.6 &56.3\\
    $+\mathcal{D}_1^1 \times 0.4$ &45.0 &51.3 &58.9 \\
    $+\mathcal{D}_1^1 \times 0.6$ &48.2 &55.8 &61.9\\
    $+\mathcal{D}_1^1 \times 0.8$ &51.1 &56.6 &63.2\\
    $+\mathcal{D}_1^1$ & 53.0 & 57.0 & 65.5 \\
    $+\mathcal{D}_1^1+\mathcal{\hat{D}}_2^1$ &58.2 &61.2 & 67.2\\
    $+\mathcal{D}_1^1+\mathcal{\hat{D}}_2^1+\mathcal{D}_3^1$ &\textbf{61.4} &\textbf{66.3} &\textbf{69.8}\\
  \hline
    \end{tabular}
    \caption{The performance of SFT with different amounts of augmented query on GSM8K.}
    \label{tab:query amount}
\end{table*}

\begin{table*}[h]
    \small
    \centering
    \begin{tabular}{lccc}
  \hline
    Response aug. & 7B & 7B-2 & 13B-2 \\
  \hline
    $\mathcal{D}$ & 35.9 & 41.6 & 50.0 \\
    \hline
    $+\mathcal{D}_1^1$ & 53.0 & 57.0 & 65.5 \\
    $+\mathcal{D}_1^1+\mathcal{D}_1^2$ &55.9 &61.4 &67.0\\
    $+\mathcal{D}_1^1+\mathcal{D}_1^2+\mathcal{D}_1^3$ & \textbf{61.3} &\textbf{64.4} &68.4\\
    $+\mathcal{D}_1^1+\mathcal{D}_1^2+\mathcal{D}_1^3+\mathcal{D}_1^4$ & 60.1 &63.8 & 69.1\\
    $+\mathcal{D}_1^1+\mathcal{D}_1^2+\mathcal{D}_1^3+\mathcal{D}_1^4+\mathcal{D}_1^5$ &60.7  &63.6  &\textbf{71.6}\\
    \hline
    $+\mathcal{D}_1^1+\mathcal{D}_1^2+\mathcal{D}_1^3$ - majority voting &56.4 &60.7 &65.7 \\
    $+\mathcal{D}_1^1+\mathcal{D}_1^2+\mathcal{D}_1^3+\mathcal{D}_1^4+\mathcal{D}_1^5$ - majority voting &58.9 &62.5 &68.3\\
  \hline
    \end{tabular}
    \caption{The performance of SFT with different amounts of augmented response on GSM8K.}
    \label{tab:response amount}
\end{table*}

\begin{table*}[h]
\centering
\subfloat[Part 1]{
\begin{tabular}{ll|cccc}
\hline
Model & Data & $\mathcal{D}$ & +$\mathcal{D}_1^1$$\times$0.2 & +$\mathcal{D}_1^1$$\times$0.4 & +$\mathcal{D}_1^1$$\times$0.6 \\
\hline
7B & training set & 56.4 & 41.8 & 51.4 & 55 \\
 & test set & 35.9 & 38.1 & 45 & 48.2 \\
7B-2 & training set & 65.2 & 48.4 & 57.4 & 64.8 \\
 & test set & 41.6 & 43.6 & 51.3 & 55.8 \\
13B-2 & training set & 75.4 & 80.4 & 80.4 & 82.6 \\
 & test set & 50 & 56.3 & 58.9 & 61.9 \\
\hline
\end{tabular}
}
\hfill
\subfloat[Part 2]{
\begin{tabular}{ll|cccc}
\hline
Model & Data & +$\mathcal{D}_1^1$$\times$0.8 & +$\mathcal{D}_1^1$ & +$\mathcal{D}_1^1+\mathcal{\hat{D}}_2^1$ &+$\mathcal{D}_1^1+\mathcal{\hat{D}}_2^1+\mathcal{D}_3^1$  \\
\hline
7B & training set & 61.6 & 71.4 & 79.8 & 83.6 \\
 & test set & 51.1 & 53 & 58.2 & 61.4 \\
7B-2 & training set & 66.6 & 79 & 85.2 & 85.6 \\
 & test set & 56.6 & 57 & 61.2 & 66.3 \\
13B-2 & training set & 82.2 & 84.4 & 86.6 & 89.2 \\
 & test set & 63.2 & 65.5 & 67.2 & 69.8 \\
\hline
\end{tabular}
}
\caption{The accuracy on the training dataset and test dataset for GSM8K.}
\label{Your label}
\end{table*}



\begin{table*}[h]
    \small
    \centering
    
    \begin{tabular}{lccc}
  \hline
    Model & 7B & 7B-2 & 13B-2  \\
    \hline
    $\mathcal{D}$ & 35.9 & 41.6 & 50.0\\
    $\mathcal{D}_1^1$ on hard  &43.0 &51.3 & 58.8\\
    $\mathcal{D}_1^1$ on medium  &43.5 &49.0 &55.6\\
    $\mathcal{D}_1^1$ on easy  &42.7 &47.6 &55.6\\
    $\mathcal{D}_1^1$ on random  &43.4 & 50.0&56.0\\
  \hline
    \end{tabular}
    \caption{The performance of SFT with query augmentation with different diffculties on GSM8K.}
    \label{tab:harder}
\end{table*}

\begin{table*}[h]
    \small
    \centering
    \begin{tabular}{lcccc}
  \hline
    Model & 7B & 7B-2 & 13B-2  \\
    \hline
    $\mathcal{D}_1^1$ on wrong &46.2 & 49.5 & 55.4\\
    $\mathcal{D}_1^1$ on random &43.6 &49.2 &54.2\\
    \hline
    \end{tabular}
    \caption{The performance of SFT with query augmentation with wrong problems or random problems on GSM8K.}
    \label{tab:wrong}
\end{table*}

\begin{table*}[h]
\centering
\begin{tabular}{|c|c|c|c|c|}
\hline
\multicolumn{4}{|c|}{closed-source models} \\
\hline
Model & \#params & GSM8K &MATH\\
\hline
GPT-4~\citep{gpt4} & - & 92.0 & 42.5\\
GPT-3.5-Turbo~\citep{instructgpt} & - & 80.8 &34.1\\
Claude-2 & - & 85.2 &32.5\\
PaLM~\citep{chowdhery2022palm} & 8B & 4.1 &1.5\\
PaLM & 62B & 33.0 &4.4\\
PaLM & 540B & 56.5 &8.8\\
PaLM-2~\citep{anil2023palm} & 540B & 80.7 &34.4\\
Flan-PaLM~\citep{anil2023palm} 2 & 540B & 84.7 &33.2\\
Minerva~\citep{Minerva} & 8B & 16.2 &14.1\\
Minerva & 62B & 52.4 &27.6\\
Minerva & 540B & 58.8 &33.6\\
\hline
\multicolumn{4}{|c|}{open-source models (1-10B)} \\
\hline
LLaMA-2~\citep{llama2} & 7B & 14.6 &2.5\\
MPT~\citep{MPT} & 7B & 6.8 &3.0\\
Falcon & 7B & 6.8 &2.3\\
Code-LLaMA~\citep{codellama} & 7B &25.2 &13.0\\
InternLM~\citep{2023internlm} & 7B & 31.2 &-\\
GPT-J~\citep{gpt-j} & 6B & 34.9 &-\\
ChatGLM-2~\citep{zeng2022glm} & 6B & 32.4 &-\\
Qwen~\citep{qwen} & 7B & 51.6 &-\\
Baichuan-2~\citep{baichuan} & 7B & 24.5 &5.6\\
MAmooTH-CoT~\citep{MAmmoTH} & 7B & 50.5 &10.4\\
RFT\cite{rft} & 7B & 50.3 & -\\
MetaMath~\citep{yu2023metamath} &7B & 66.5&19.8\\
\textbf{MuggleMath-7B}& 7B & \textbf{69.8 }&\textbf{25.8}\\

\hline
\multicolumn{4}{|c|}{open-source models (11-50B)} \\
\hline
LLaMA-2~\citep{llama2} & 13B & 28.7 & 3.9\\
Platypus~\citep{Platypus}  & 13B & 25.7 &2.5\\
LLaMA-2 & 34B & 42.2 &6.2\\
MPT~\citep{MPT} & 30B & 15.2 &3.1\\
Falcon~\citep{falcon} & 40B & 19.6 &2.5\\
Vicuna~\citep{vicuna} & 13B & 27.6 &-\\
Baichuan-2~\citep{baichuan} & 13B & 52.8 &10.1\\
MAmooTH-CoT~\citep{MAmmoTH} & 13B & 56.3 &12.9\\
Code-LLaMA~\citep{codellama} & 13B &36.1 &16.4\\
RFT\cite{rft} & 13B & 54.8 &-\\
WizardMath~\citep{luo2023wizardmath} & 13B & 63.9 &14.0 \\
MetaMath~\citep{yu2023metamath} & 13B& 72.3&22.4\\
\textbf{MuggleMath-13B} & 13B & \textbf{74.3} &\textbf{30.7}\\
\hline
\multicolumn{4}{|c|}{open-source models (51-70B)} \\
\hline
LLaMA-2~\citep{llama2} & 70B & 56.8 &13.5\\
RFT~\citep{rft} & 70B & 64.8 &-\\
Platypus~\citep{Platypus}  & 70B & 70.6 &15.6\\
MAmooTH-CoT~\citep{MAmmoTH} & 70B & 71.4 &21.1\\
WizardMath\cite{luo2023wizardmath} & 70B & 81.6 &22.7\\
MetaMath~\citep{yu2023metamath} &70B & 82.3&26.6\\
\textbf{MuggleMath-70B} & 70B & \textbf{82.7 }&\textbf{36.3}\\
\hline
\end{tabular}
\caption{Model comparison of MuggleMath and  a broad range of state-of-the-art approaches. 
}
\label{More comparsion}
\end{table*}

\begin{table*}[h!]
  \centering
  \begin{tabular}{|l|c|c|c|}
    \hline
    & \textbf{7B} & \textbf{7B-2} & \textbf{13B-2} \\ \hline
    \multicolumn{4}{|c|}{\textbf{Change-Test}} \\ \hline
    SFT & 26.2 & 30.1 & 38.6 \\ \hline
    MuggleMath & 60.1 & 62.8 & 67.1 \\ \hline
    \multicolumn{4}{|c|}{\textbf{Aug-Test}} \\ \hline
    SFT & 14.2 & 17.2 & 22.4 \\ \hline
    MuggleMath & 40.1 & 44.3 & 49.3 \\ \hline
  \end{tabular}
    \caption{We have perturbed two new test sets based on the original GSM8K test set.
(A) Change-Test, is created by altering the numerical values in the GSM8K test set questions and correspondingly modifying the answers. There are 1211 query-response pairs in the Change-Test.
(B) Aug-Test, is generated by augmenting the test set in the same manner as we did for the training set. There are 1378 query-response pairs in the Aug-Test.}
\label{perturb_test}
\end{table*}

\begin{table*}[h!]
\centering
\begin{tabular}{lcccc}
\toprule
Subject & MATH & GSM8k & GSM8K+$\mathcal{D}_1^i$ & GSM8K+$\sum_{i=1}^{3}{\mathcal{D}_1^i}+\mathcal{D}_2^1+\mathcal{D}_3^1$\\
\midrule
Counting \& Probability & 10.5 & 13.2 & 7.9 & 5.3 \\
Algebra & 7.3 & 12.1 & 12.9 & 16.9 \\
Prealgebra & 8.5 & 13.4 & 8.5 & 11.0 \\
Geometry & 2.4 & 9.8 & 4.9 & 2.4 \\
Intermediate Algebra & 6.2 & 5.2 & 3.1 & 5.2 \\
Number Theory & 3.2 & 6.5 & 6.5 & 8.1 \\
Precalculus & 3.6 & 5.4 & 7.1 & 7.1 \\
\bottomrule
\end{tabular}
\caption{Transfer learning accuracy on subsets of MATH  for LLaMA-13B-2}
\label{subset1}
\end{table*}

\begin{table*}[h]
\centering
\begin{tabular}{lcccc}
\toprule
Subject & MATH & GSM8k & GSM8K+$\mathcal{D}_1^i$ & GSM8K+$\sum_{i=1}^{3}{\mathcal{D}_1^i}+\mathcal{D}_2^1+\mathcal{D}_3^1$\\
\midrule
Prealgebra              & 12.2 &  9.8 & 11.0 & 12.2 \\
Number Theory           &  6.5 &  9.7 &  6.5 &  9.7 \\
Algebra                 &  7.3 &  5.6 &  5.6 & 15.3 \\
Intermediate Algebra    &  2.1 &  4.1 &  4.1 &  1.0 \\
Precalculus             &  3.6 &  1.8 &  3.6 &  1.8 \\
Counting \& Probability &  5.3 &  7.9 &  5.3 & 13.2 \\
Geometry                &  -   &  2.4 &  -   &  -   \\
\bottomrule
\end{tabular}
\caption{Transfer learning accuracy on subsets of MATH  for LLaMA-7B-2}
\label{subset2}
\end{table*}

\begin{table*}[h]
\centering
\begin{tabular}{lcccc}
\toprule
Subject & MATH & GSM8k & GSM8K+$\mathcal{D}_1^i$ & GSM8K+$\sum_{i=1}^{3}{\mathcal{D}_1^i}+\mathcal{D}_2^1+\mathcal{D}_3^1$\\
\midrule
Prealgebra              & 7.3 &  9.8 &  7.3 & 14.6 \\
Number Theory           & 6.5 &  3.2 &  1.6 &  3.2 \\
Algebra                 & 6.5 &  4.8 & 11.3 &  4.8 \\
Intermediate Algebra    & 3.1 &  1.0 &  5.2 &  3.1 \\
Precalculus             & -   &  3.6 &  5.4 &  1.8 \\
Counting \& Probability & 2.6 &  2.6 &  2.6 &  7.9 \\
Geometry                & 4.9 &  4.9 &  2.4 &  2.4 \\
\bottomrule
\end{tabular}
\caption{Transfer learning accuracy on subsets of MATH  for LLaMA-7B}
\label{subset3}
\end{table*}

\begin{table*}[h]
\centering
\begin{tabular}{lccc}
\toprule
Subject & MATH & GSM8k & GSM8K+$\mathcal{D}_1^1$  \\
\midrule
Prealgebra              & 8.5 & 12.2 & 14.6 \\
Number Theory           & 3.2 & 12.9 & 3.2 \\
Algebra                 & 7.3 & 8.1 & 12.1 \\
Intermediate Algebra    & 6.2 & 5.2 & 7.2 \\
Precalculus             & 3.6 & - & - \\
Counting \& Probability & 10.5 & 5.3 & 10.5 \\
Geometry                & 2.4 & 7.3 & 4.9 \\
\bottomrule
\end{tabular}
\caption{Multi-task learning accuracy on subsets of MATH  for LLaMA-13B-2}
\label{subset4}
\end{table*}

\begin{table*}[h]
\centering
\begin{tabular}{lccc}
\toprule
Subject & MATH & GSM8k & GSM8K+$\mathcal{D}_1^1$  \\
\midrule
Prealgebra              & 12.2 & 11.0 &  4.9 \\
Number Theory           &  6.5 &  6.5 &  3.2 \\
Algebra                 &  7.3 &  5.6 & 11.3 \\
Intermediate Algebra    &  2.1 &  6.2 &  -   \\
Precalculus             &  3.6 &  -   &  -   \\
Counting \& Probability &  5.3 &  5.3 &  2.6 \\
Geometry                &  -   &  7.3 &  7.3 \\
\bottomrule
\end{tabular}
\caption{Multi-task learning accuracy on subsets of MATH  for LLaMA-7B-2}
\label{subset5}
\end{table*}

\begin{table*}[h]
\centering
\begin{tabular}{lccc}
\toprule
Subject & MATH & GSM8k & GSM8K+$\mathcal{D}_1^1$  \\
\midrule
Prealgebra              & 7.3 & 6.1 & 7.3 \\
Number Theory           & 6.5 & 3.2 & 3.2 \\
Algebra                 & 6.5 & 6.5 & 10.5 \\
Intermediate Algebra    & 3.1 & 2.1 & - \\
Precalculus             & - & 1.8 & - \\
Counting \& Probability & 2.6 & 2.6 & 5.3 \\
Geometry                & 4.9 & 9.8 & 2.4 \\
\bottomrule
\end{tabular}
\caption{Multi-task learning accuracy on subsets of MATH  for LLaMA-7B}
\label{subset6}
\end{table*}

\end{document}